\documentclass[runningheads]{llncs}

\usepackage{eccv}

\usepackage{eccvabbrv}

\usepackage{graphicx}
\usepackage{booktabs}

\usepackage[accsupp]{axessibility}  

\usepackage{hyperref}

\usepackage{orcidlink}

\usepackage[normalem]{ulem}
\PassOptionsToPackage{usenames, dvipsnames}{color}
\usepackage{xcolor,colortbl}
\usepackage{ifthen}
\usepackage{xspace}
\usepackage{amsmath}
\usepackage{comment}
\usepackage{enumitem}
\usepackage{subcaption}
\usepackage{float}

\usepackage{multirow}
\usepackage{tabularx,array}
\newcolumntype{Y}{>{\centering\arraybackslash}X}
\usepackage{booktabs,makecell,xcolor}
\definecolor{rowgray}{RGB}{245,245,245}

\newcommand{\colmap}{COLMAP\xspace}
\newcommand{\glomap}{GLOMAP\xspace}
\newcommand{\theia}{Theia\xspace}

\newcommand{\resfm}{RESfM\xspace}
\newcommand{\vggsfm}{VGGSfM\xspace}
\newcommand{\vggt}{VGGT\xspace}
\newcommand{\method}{VGPA\xspace}
\usepackage{rotating}   

\usepackage{graphicx}
\usepackage{capt-of} 
\usepackage{booktabs} 
\usepackage{multirow} 
\usepackage{siunitx} 

\definecolor{rowgray}{gray}{0.9}
\newcolumntype{d}[1]{S[table-format=#1]}
  \usepackage{adjustbox}

\definecolor{bestDeep}{rgb}{1, 1, 0.2}  
\definecolor{bestClass}{rgb}{0.7, 0.8, 0.65}

\definecolor{tabfirst}{rgb}{0.4, 0.65, 0.3} 
\definecolor{tabsecond}{rgb}{0.7, 0.8, 0.65} 

\definecolor{Gray}{gray}{0.3}
\definecolor{LightCyan}{rgb}{0.88,1,1}
\newcolumntype{a}{:c|}
\newcolumntype{d}{:c}

\begin{document}

\title{Learning Global Camera Poses from Noisy View-Graphs for Structure from Motion}

\titlerunning{Learning Global Camera Poses from Noisy View-Graphs for SfM}

\author{Fadi Khatib\orcidlink{0009-0008-8215-1334} \and
Meirav Galun\orcidlink{0009-0006-1659-7093} \and
Ronen Basri\orcidlink{0000-0001-8053-2151}}

\authorrunning{F. Khatib et al.}

\institute{Weizmann Institute of Science\\ 
\noindent\textbf{Project webpage:} \href{https://vgpa-sfm.github.io/}{vgpa-sfm.github.io}}

\maketitle

\begin{abstract}

Camera pose estimation is a key step in 3D reconstruction and view-synthesis pipelines. We present a deep, global Structure-from-Motion framework based on learned view-graph aggregation. Our method employs a permutation-equivariant, edge-conditioned graph neural network that takes noisy pairwise relative poses as input and outputs globally consistent camera extrinsics. The network is trained without ground-truth supervision, relying solely on a relative-pose consistency objective. This is followed by 3D point triangulation and robust bundle adjustment. Our approach is efficient, scalable to more than a thousand images, and robust to graph density. We evaluate our method on MegaDepth, 1DSfM, Strecha, and BlendedMVS. These experiments demonstrate that our method achieves superior rotation and translation accuracy compared to deep track-centric methods while registering more images across many scenes, and competitive results compared to state-of-the-art classical pipelines, while being much faster.

\end{abstract}

\section{Introduction}
\label{sec:intro}

Camera pose recovery is an essential part of 3D scene reconstruction and view synthesis applications. Many common Multiview Stereo (MVS) \cite{seitz2006comparison, yao2018mvsnet} and view synthesis methods, including Neural Radiance Fields (NeRF) \cite{mildenhall2021nerf} and Gaussian Splatting (GS) \cite{kerbl20233d} rely on accurate camera poses computed in preprocessing. View synthesis methods, in particular, have gained much popularity in recent years, as they can produce novel, realistically looking images and walkthroughs for complex scenes.

Multiview Structure-from-Motion (SfM) techniques provide reliable tools for camera pose recovery. Sequential pipelines, e.g., COLMAP \cite{schoenberger2016sfm}, solve for one camera at a time, enriching the recovered set of camera poses and 3D points by processing image by image. These, generally highly accurate techniques, are relatively slow when applied to large collections of images, and their performance depends on the order in which the images are processed. Projective factorization techniques \cite{sturm1996factorization} simultaneously solve for all cameras and point tracks. These methods, however, attempt to factor large tensors that include all the track points.

In the past decade, \emph{global methods} emerged as an alternative to sequential and factorization methods. Global methods use a technique called \emph{motion averaging}; given pairwise relative camera motion measurements, they seek to recover the location and orientation (and possibly also the intrinsic parameters) of cameras in a global coordinate system. Typically, this is done by solving separately for rotations and translations \cite{moulon2016openmvg, sweeney2015theia}, while some recent works developed techniques for directly averaging essential and fundamental matrices \cite{kasten2019algebraic, kasten2019gpsfm}. Global methods can be more efficient than both sequential and factorization-based techniques, as they only solve for pose and therefore do not need to access and manipulate point tracks, except in the final bundle adjustment (BA) step.

In this paper, we reexamine the use of global SfM through the lens of \emph{learned view-graph aggregation}. Specifically, we propose an efficient permutation-equivariant, edge-conditioned graph neural network (GNN) that takes as input noisy estimates of pairwise relative camera poses associated with the edges of a view graph, and outputs globally consistent camera extrinsics. The network is trained \emph{without ground-truth supervision} using only a relative-pose consistency objective. Unlike existing deep-based approaches to SfM \cite{khatib2025resfm, moran2021deep, brynte2023learning}, our pose regression network does not use point tracks; it does not predict 3D points and does not rely on a reprojection loss. At test time, we use our network to predict global camera poses. Then, we improve our camera pose predictions by triangulating point tracks and applying robust BA. Finally, an optional and efficient \emph{view reintegration} step is applied to recover cameras that were discarded in the process by the pipeline, increasing camera coverage.

\begin{figure*}[t]
  \centering
  \captionsetup[sub]{font=small,justification=centering}

  \begin{subfigure}[t]{0.48\linewidth}
    \centering
    \includegraphics[height=3.6cm, trim=260 10 170 10, clip]{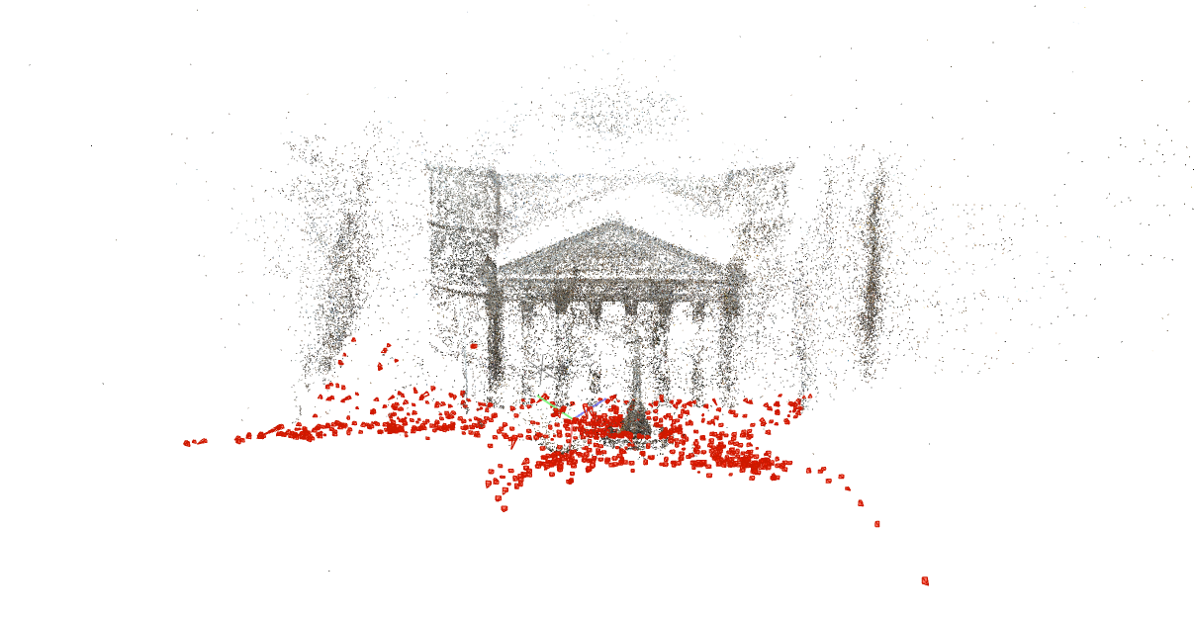}
    \caption{Scene 0023}
    \label{fig:reconstruction1_b}
  \end{subfigure}\hfill
  \begin{subfigure}[t]{0.48\linewidth}
    \centering
    \includegraphics[height=3.6cm, trim=260 10 170 30, clip]{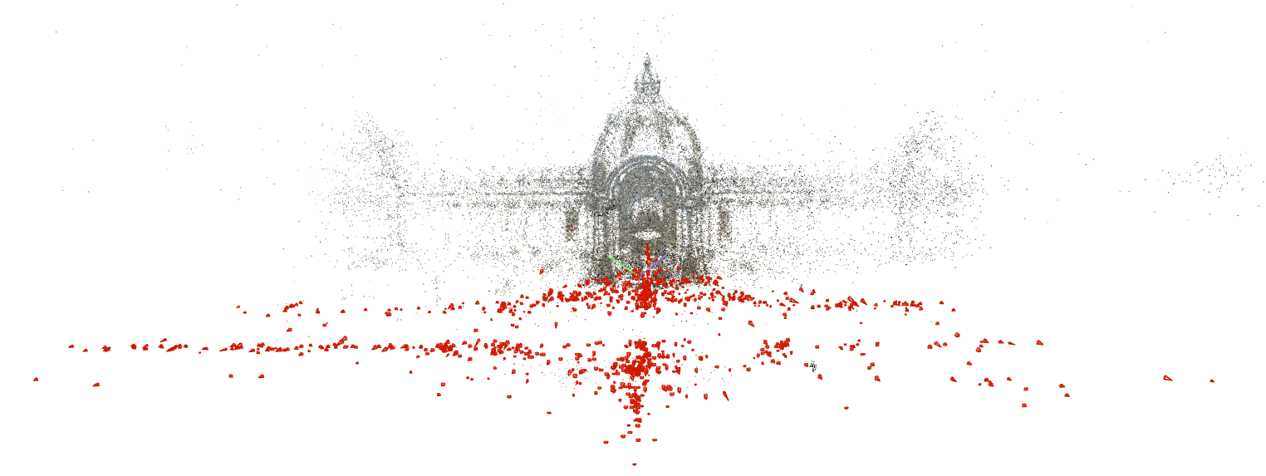}
    \caption{Scene 0455}
    \label{fig:reconstruction2_b}
  \end{subfigure}

  \vspace{-4pt}
  \caption{3D reconstructions and recovered camera parameters produced by \method{} on two large scenes ($N_c\!>\!1000$ images). \method{} registers almost all images and scales to thousand-image collections, in contrast to existing image-based deep methods (e.g., \vggt{}, \vggsfm{}).}
  \label{fig:large_scenes_fig}
\end{figure*}

Our approach is efficient and achieves high accuracy. It copes well with large-scale inputs, including scenes with more than a thousand images. In addition, our method is not restricted to exhaustive pairwise matching. We also evaluate a sparse retrieval-based view graph constructed from the top-30 MegaLoc~\cite{berton2025megaloc} neighbors, and observe that VGPA often preserves strong registration coverage and accuracy despite the large reduction in edge density.
We note that in the uncalibrated setting, we optimize jointly for the intrinsics and extrinsics parameters during BA. 

We perform an extensive experimental evaluation on challenging datasets, including MegaDepth and 1DSfM. These experiments demonstrate that our learned pose averaging achieves lower camera position and orientation errors than existing deep track-centric methods while registering more images on many scenes \cite{khatib2025resfm, moran2021deep, brynte2023learning}, and is competitive with strong classical pipelines. Similar results are obtained on smaller calibrated benchmarks for which ground truth measurements are available (Strecha and BlendedMVS) and on scenes containing challenging cyclic trajectories, where reprojection-centric methods such as \cite{khatib2025resfm, moran2021deep, brynte2023learning} often struggle.

Below we summarize our contributions.

\begin{itemize}
\item We introduce \textbf{\method{}}, a permutation-equivariant GNN that performs \emph{\textbf{V}iew-\textbf{G}raph \textbf{P}ose \textbf{A}veraging}, learning to aggregate noisy pairwise relative poses to recover globally consistent camera extrinsics.

\item \method{} achieves accurate camera pose and structure recovery, matching the accuracy of state-of-the-art classical SfM pipelines while \textbf{being significantly faster}, and substantially outperforming recent feed-forward deep reconstruction methods on large-scale scenes.

\item We train \method{} in a \textbf{self-supervised manner} by enforcing relative-pose consistency only, while 3D structure is recovered through triangulation followed by robust bundle adjustment.

\item We show that \method{} can operate on sparse retrieval-based view graphs, obtaining competitive results when exhaustive pairwise matching is replaced by a top-$k$ retrieval graph, despite large differences in edge density.

\end{itemize}

\section{Related work}
\label{sec:related work}

A popular classical method for Structure-from-Motion (SfM) uses an incremental algorithm in which images are processed one at a time, gradually extending the recovered set of camera poses and 3D structure. \cite{agarwal2011building,schoenberger2016sfm, snavely2006photo, wu2013towards}. While these methods achieve highly accurate reconstruction, they are inefficient when applied to large image collections, and their results depend on the order in which images are processed.

A second approach uses projective factorization to solve simultaneously for camera pose and 3D structure on all input images \cite{sturm1996factorization,dai2010element,lin2017factorization}. This method uses the observation that point track matrices are rank 4 when the points are scaled properly. Classical algorithms based on SVD factorization, however, are restricted to uncalibrated settings and do not handle missing data or outliers. Inspired by these techniques, several recent works train equivariant network architectures to jointly estimate camera poses and 3D structure from point tracks \cite{moran2021deep,brynte2023learning,chen2024deepaat,khatib2025resfm}. These methods use either set-of-sets or graph transformer network architectures and are trained with either supervised or unsupervised data. An inlier/outlier classifier is incorporated for improved robustness~\cite{khatib2025resfm}. Accurate pose recovery results were achieved with this method. However, it tends to over-prune valid inliers, leading to occasional registration failures and reduced image coverage.

Our method follows a third approach, commonly referred to as a \emph{global approach}. Global methods handle all images simultaneously by applying manifold averaging to ensure the consistency of pairwise pose relations (rotations and translations) inferred from the essential matrices. Existing methods commonly solve first for camera orientations, and next for location and scales \cite{martinec2007robust, ozyecsil2017survey,sweeney2015theia, moulon2016openmvg}. \cite{kasten2019algebraic, kasten2019gpsfm} introduced an averaging method for averaging essential and fundamental matrices, solving for all of these parameters in a single optimization. With the exception of \cite{pan2024global}, these methods require a separate step of 3D point triangulation. Theia \cite{sweeney2015theia} and the recent GLOMAP \cite{pan2024global}, in particular, were shown to yield accurate recovery.

Several recent works train networks to solve rotation averaging on the view graph. NeurORA \cite{purkait2020neurora} learns to denoise pairwise relative rotations and aggregates them to recover global orientations, while \cite{li2021pogo} applies message passing on pose graphs to iteratively update node rotations. These methods only address rotation averaging; they are trained on supervised data and tested in limited settings that do not include cross-dataset generalization. In contrast, our method is trained with unsupervised data and recovers the full camera extrinsics.

Recent learnable SfM methods such as VGGSfM \cite{wang2023visual}, DUST3R \cite{wang2023dust3r}, and MAST3R \cite{leroy2024grounding} are restricted to processing only a small number of input images, whereas Ace-Zero \cite{brachmann2024scene} and FlowMap \cite{smith2024flowmap} are tailored for video sequences under constant illumination.  More recently, \vggt \cite{wang2025vggt} introduced an end-to-end transformer that jointly predicts camera poses, dense 3D structure, and point tracks. Although promising, \vggt requires substantial supervised training and is currently restricted to images on the order of $\sim$ 200.\footnote{In May 2026, after our experiments were completed, the VGGT authors reported an implementation fix that removes redundant intermediate tensors kept in memory, allowing the model to process roughly $2$-$3\times$ more frames under the same GPU memory budget.} Fast3R \cite{yang2025fast3r} scales to larger collections but typically attains lower accuracy than \vggt{} at comparable settings. Continuous 3D perception methods such as CUT3R \cite{wang2025continuous} maintain a persistent scene state and incrementally predict camera parameters and pointmaps from an image stream, while TTT3R \cite{chen2025ttt3r} extends this framework with test-time training to improve long-sequence reconstruction without modifying the model parameters. However, while such methods scale to long sequences, they perform poorly on large unordered image collections.

In this paper, we introduce a learned view-graph pose averaging module implemented with a permutation-equivariant graph neural network. Trained without ground-truth supervision, our method achieves competitive accuracies at lower runtime than strong global SfM baselines and surpasses prior deep factorization approaches in both accuracy and camera coverage.

\begin{figure*}{}
    \centering
    \includegraphics[width=0.9\textwidth, trim={0cm, 0cm, 2cm, 0cm}, clip]{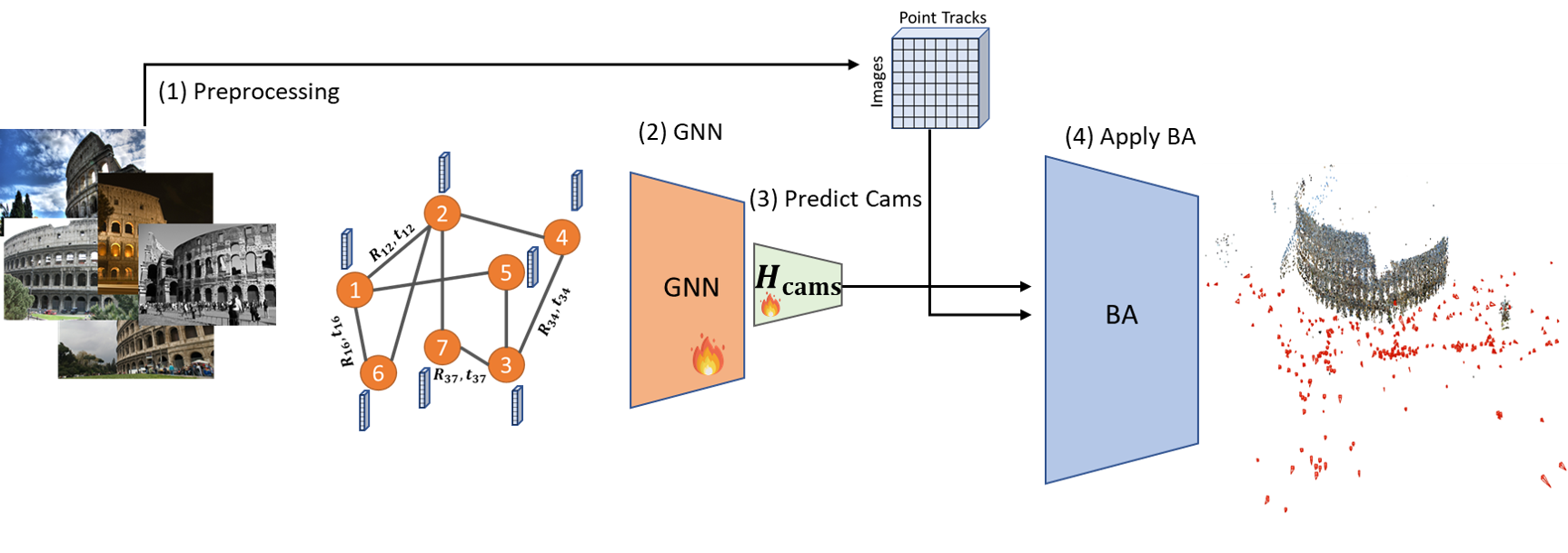}
    \caption{{\small \textbf{Method overview.} (1) \emph{Preprocessing:} estimate pairwise relative poses from essential matrices and build the view graph; extract point tracks. (2) \emph{GNN:} a permutation-equivariant, edge-conditioned GNN aggregates the view graph to produce camera embeddings. (3) \emph{Predict cams:} a small head $H_{\text{cams}}$ regresses global extrinsics $(R_i,\mathbf t_i)$ from the embeddings. (4) \emph{Triangulation + BA:} using the predicted cameras and the point tracks, we triangulate 3D points and run robust bundle adjustment.}} \label{fig:network_architecture}
\end{figure*}
\section{Method}
\label{sec:method}

Given a collection of $m$ images of a stationary scene, we assume, as in standard SfM pipelines, that in preprocessing we extract (1) essential matrices and (2) a collection of point tracks, which will form the input to our pipeline. Our objective is to recover the camera matrices for all the given images and a triangulated 3D location for each track. Below, we describe each step in our method. 

\subsection{Preprocessing} \label{sec:preprocessing}

Denote our input images by $I_1,...,I_m$. Following standard SfM pipelines, we begin by detecting and matching features across the images using
SIFT \cite{lowe2004distinctive}. We next apply RANSAC \cite{bolles1981ransac} and obtain a partial collection of pairwise essential matrices $\{E_{ij}\}_{i,j \in [m]}$, denoted by $\mathcal{E}$.  Each essential matrix encodes the relative rotation $R_{ij}$ and translation ${\bf t}_{ij}$ between camera $P_i$ and $P_j$. We extract the rotation and translation by decomposing the essential matrix, while enforcing positive depth. Note that ${\bf t}_{ij}$ is determined at this point only up to scale. These pairwise rotation and translation measurements serve as input to our pose averaging module.

A second outcome of the procedure above comprises pairs of matched feature points across images. We next use heuristics (as in, e.g., \cite{schoenberger2016sfm}) to join such pairs to form longer tracks. Each track is a set $T_k=\{{\bf x}_{i_1,k},{\bf x}_{i_2,k},...\}$ with $i_1,i_2,... \in [m]$, and we assume that $T_k$ contains the projected locations of a single 3D scene point, denoted ${\bf X}_k$, onto $I_{i_1}, I_{i_2}, ...$. These tracks are generally contaminated by small displacement errors (noisy measurements) and outliers.

We will use this collection of point tracks at a later stage to triangulate the 3D structure using the predicted absolute camera poses.

\subsection{Network architecture}

Our network applies \emph{pose averaging} to the view graph. As is shown in ~\cref{fig:network_architecture}, it comprises two modules: (i) a permutation-equivariant, edge-conditioned GNN that aggregates pairwise relative poses into camera embeddings; and (ii) a regression head that predicts global camera parameters from these embeddings.
  \noindent\textbf{Pose-averaging GNN.}
  We build a viewing graph $\mathcal{G}=(\mathcal{V},\mathcal{E})$ whose nodes index the $m$ images and whose edges carry relative-pose measurements. For each
  edge $(i,j)\in\mathcal{E}$ we define
  \[
  \mathbf e_{ij}=\phi_e\!\big(\log R_{ij}^{\text{RANSAC}},\,\mathbf t_{ij}^{\text{RANSAC}}\big),
  \]
  where $\log R_{ij}^{\text{RANSAC}}\in\mathbb{R}^3$ is the SO(3) log-map of the relative rotation recovered from RANSAC and $\mathbf
  t_{ij}^{\text{RANSAC}}\!\in\!\mathbb S^2$ is the unit translation direction recovered from the essential matrix, and $\phi_e$ is a learned MLP. Since the available geometric information
  resides on the edges of the view graph, node embeddings are initialized with a shared learnable token $\mathbf{h}_i^{(0)} = \mathbf{h}_0$.

  We apply edge-conditioned message passing with degree-normalized mean aggregation:
  \begin{align*}
  \tilde{\mathbf m}_i^{(\ell)} &= \sum_{j\in\mathcal{N}(i)}
  \phi_m\!\big(\mathbf h_i^{(\ell)},\,\mathbf h_j^{(\ell)},\,\mathbf e_{ij}\big),\\
  \mathbf m_i^{(\ell)} &= \frac{1}{|\mathcal{N}(i)|}\,\tilde{\mathbf m}_i^{(\ell)},\\
  \mathbf h_i^{(\ell+1)} &= \mathrm{LN}\!\Big(\mathbf h_i^{(\ell)} + \mathrm{Drop}\big(\psi\big(\mathrm{LN}(\mathbf h_i^{(\ell)}),\, \mathbf
  m_i^{(\ell)}\big)\big)\Big),
  \end{align*}
  for $\ell=0,\dots,L-1$, where $\phi_m$ and $\psi$ are MLPs, $\mathrm{LN}$ denotes Layer Normalization~\cite{ba2016layer}, and $\mathrm{Drop}$ denotes
  Dropout~\cite{srivastava2014dropout}. The network is equivariant to node relabelings (permutations) of $\mathcal{G}$. In ablation experiments we also evaluated
  replacing the message passing operator with a graph attention mechanism (GATv2), but observed no performance improvements while incurring higher computational
  cost. The final node embeddings are $z_i =\mathbf h_i^{(L)}$.

  \noindent\textbf{Pose regression head.}
  The pose regression head obtains as input the per-camera embeddings $\mathbf z_i$ produced by the pose-averaging GNN. A 3-layer MLP head $H_{\mathrm{cams}}$
  maps these embeddings to camera parameters,
  \[
  (\mathbf t_i,\mathbf{q}_i)=H_{\mathrm{cams}}(\mathbf z_i), \qquad \mathbf{\tilde{q}}_i \leftarrow \mathbf q_i/\|\mathbf q_i\|,
  \]
  where $\mathbf t_i\in\mathbb{R}^3$ and $\mathbf{\tilde{q}}_i\in\mathbb{H}$ is a unit quaternion.

\subsection{Output and loss}
  Our network predicts the $m$ internally calibrated cameras $P_1,\dots,P_m$. Each camera is parameterized as $P_i=[R_i \mid \mathbf t_i]$ with $R_i\in SO(3)$
  and $\mathbf t_i\in\mathbb{R}^3$; the camera center is $-R_i^\top \mathbf t_i$.

  Training is unsupervised and seeks cameras $P_1,\dots,P_m$ that best agree with the pairwise relative-pose estimates. We therefore minimize a relative-pose
  consistency objective. Specifically, we use
  \begin{align}
      && \mathcal{L}_{\text{RelPose}} = \,\frac{1}{|\mathcal{E}|} \! \sum_{(i,j)\in\mathcal{E}} \! d_R \! \big( \hat{R}_{ij}, \, R_{ij}^{\text{RANSAC}} \big)
      + \, \frac{1}{|\mathcal{E}|}\!\sum_{(i,j)\in\mathcal{E}}\! d_t\!\big(\hat{\mathbf{t}}_{ij}, \,\mathbf{t}_{ij}^{\text{RANSAC}} \big),
  \label{eq:loss_relpose}
  \end{align}
  where $\hat R_{ij}$ and $\hat {\bf t}_{ij}$ denote the relative rotation and translation estimated from the output cameras $P_i$ and $P_j$ using
  \begin{equation}
  \label{eq:relpose}
  \hat R_{ij} \;=\; R_j R_i^\top,
  \qquad
  \mathbf{\hat{t}}_{ij} \;=\; \mathbf{t}_j - R_j R_i^\top \mathbf{t}_i,
  \end{equation}
  $R_{ij}^{\text{RANSAC}}$ and ${\bf t}_{ij}^{\text{RANSAC}}$ are the corresponding rotation and translation obtained with RANSAC in preprocessing,
  $d_R(R_1,R_2)=\arccos\left(\frac{\text{trace}(R_1^\top R_2)-1}{2}\right)$ is the geodesic rotation error, and $d_t(\mathbf{a},\mathbf{b})=\arccos\langle
  \mathbf{a},\mathbf{b}\rangle$ measures directional disagreement.


\noindent\textbf{Training protocol.}
We iterate over all training scenes in each epoch.
Models are selected by early stopping on a held-out validation set; we report the checkpoint with the lowest validation error. Additional implementation details and architectural specifications are provided in the Supplementary Material.

\noindent\textbf{Inference.}
On an \emph{unseen} scene, the model predicts all camera poses in a single forward pass. We then fine-tune on the target scene with the unsupervised objective (no ground-truth labels) for $T_{\text{ft}}\!=\!200$ steps. Next, we triangulate using DLT \cite{hartley2003multiple} to recover 3D point positions from the estimated camera poses and point tracks, and finally perform a robust bundle adjustment initialized with the camera poses predicted by the network and the triangulated points.

\section{Experiments}
\label{sec:experiments}

\subsection{Datasets}

We train our network on scenes from the MegaDepth dataset \cite{li2018megadepth} and then test it on a diverse range of real-world scenes that include novel scenes from the MegaDepth dataset as well as cross-dataset generalization tests on the 1DSfM dataset \cite{wilson2014robust}, Strecha~\cite{strecha2008benchmarking}, and BlendedMVS~\cite{yao2020blendedmvs}. We refer the reader to the supplementary material for hyperparameters and further technical details.

\noindent\textbf{MegaDepth~\cite{li2018megadepth}.} The MegaDepth dataset includes 196 different outdoor landmark scenes curated from the internet. We followed the train/test split as in~\cite{khatib2025resfm}, including subsampling of scenes with more than 1000 images. In Table~\ref{tab:megadepth_mean_errors}, above the middle rule are scenes with fewer than 1000 images, while the scenes below the rule are subsampled.

\noindent\textbf{1DSFM~\cite{wilson2014robust}.} 1DSFM is a collection of diverse urban scenes reconstructed from community photo collections. We use this dataset to test our method (trained on the MegaDepth dataset) in cross-dataset generalization experiments, demonstrating large-scale reconstructions in realistic settings.

\noindent\textbf{Strecha~\cite{strecha2008benchmarking}.} The Strecha dataset consists of small outdoor scenes ($\le 30$ images) and includes ground-truth data acquired with a LIDAR system. 

\noindent\textbf{BlendedMVS~\cite{yao2020blendedmvs}.} The BlendedMVS dataset includes synthetic scenes with textured meshes rendered and blended to produce color images and depth maps, providing ground truth camera poses.

\noindent\textbf{Ground truth camera poses.}
Many challenging datasets, including MegaDepth and 1DSFM, lack ground-truth measurements; therefore, as is common in the field, we use camera poses computed with COLMAP~\cite{schoenberger2016sfm}, a state-of-the-art incremental Structure from Motion (SfM) method, to generate ``ground truth" camera poses. COLMAP is widely used for this purpose (see ~\cite{jiang2013global,wilson2014robust,cui2015global,ozyesil2015robust,brynte2023learning,khatib2025resfm,zhang2024raydiffusion}) due to its accurate and robust performance. To evaluate our method with real ground truth, we additionally show results on the smaller datasets Strecha~\cite{strecha2008benchmarking} and BlendedMVS~\cite{yao2020blendedmvs}.

\vspace{-3pt}
\subsection{Baselines}
\label{subsec:Baselines}

With the exception of \vggt \cite{wang2025vggt}, the settings and results for all baselines below were taken from \cite{khatib2025resfm}.

\noindent\textbf{RESfM \cite{khatib2025resfm}.}
RESfM is a robust deep equivariant SfM model that operates on a point-track tensor using a sets-of-sets permutation-equivariant architecture. It augments prior equivariant factorization by adding a multiview inlier/outlier classifier integrated into the same equivariant backbone and concludes with a robust bundle-adjustment stage.

\noindent\textbf{Theia \cite{sweeney2015theia}.} A global SfM pipeline that applies rotation averaging, followed by translation averaging, and finally 3D point triangulation.

\noindent\textbf{\glomap \cite{pan2024global}.} A global SfM pipeline that first applies rotation averaging, followed by an integrated step of translation averaging and point triangulation. 

\noindent\textbf{VGGSfM \cite{wang2024vggsfm}} is a differentiable, trainable SfM pipeline. 

\noindent\textbf{MASt3R \cite{leroy2024grounding}.} An SfM pipeline that utilizes a global alignment procedure to merge pairwise pointmap predictions.

\noindent\textbf{\vggt \cite{wang2025vggt}.} \vggt is a feed-forward, end-to-end multi-view transformer network that jointly predicts cameras, depth, point maps, and tracks for up to about 200 views. It uses alternating inter-frame/global attention and is additionally refined with BA.

We also evaluate several recent feed-forward 3D reconstruction models designed to scale to large image collections, including \textbf{FAST3R} \cite{yang2025fast3r}, a transformer-based model that predicts per-pixel 3D locations in a shared reference frame in a single forward pass, and \textbf{CUT3R} and \textbf{TTT3R} \cite{wang2025continuous,chen2025ttt3r}, continuous 3D perception models that incrementally reconstruct scene geometry and camera parameters from an image stream while maintaining a persistent scene representation.



\subsection{Metrics and evaluation}

To evaluate our results, we first align the predicted scenes to the ground truth by applying a per-scene 3D similarity transformation. We then compare our camera orientation predictions with the ground truth ones using angular differences in degrees. We measure differences between our predicted and ground truth camera locations using the $l_2$ distance. For a fair comparison, both our method and all baseline methods operating on point tracks were evaluated using the same set of point tracks. For all methods, we apply a final post-processing step of robust bundle adjustment. Additional tables reporting AUC metrics are provided in the supplementary.

\vspace{-1pt}
\begin{table*}[t]

\caption{{\small {\bf MegaDepth experiment.} For each scene, we show the number of input images (denoted $N_c$) and the fraction of outliers. For each model, we show the number of images used for reconstruction (denoted $N_r$) and mean values of the rotation error (in degrees) and translation error. Above the middle rule are Group 1 scenes with $<1000$ images; below are Group 2 scenes with $>1000$ images, subsampled to 300 for testing. Winning results are marked in \textbf{\underline{bold and underlined}}.}}

    \label{tab:megadepth_mean_errors}
    
    \centering
    \tiny 
    \rowcolors{3}{rowgray}{white} 
    \setlength{\tabcolsep}{3pt} 
    \renewcommand{\arraystretch}{1.1} 
    \begin{adjustbox}{max width=1.0\textwidth}
    \begin{tabular}{ccc|ccc|ccc||ccc|ccc}
    
\multirow{2}{*}{ Scene } &
\multirow{2}{*}{ $N_c$ } &
\multirow{2}{*}{ Outliers\% } &
\multicolumn{3}{c}{\textbf{Ours}}&
\multicolumn{3}{c}{RESfM} &
\multicolumn{3}{c}{\theia}&
\multicolumn{3}{c}{\glomap}\\

 &  &  & $N_r$ & Rot  & Trans   & $N_r$ & Rot  & Trans   &$N_r$ & Rot  & Trans    & $N_r$ & Rot  & Trans \\
 
0238 & 522 & $ 44.6\% $  & 486  & 1.06  & $\mathbf{\underline{0.285}}$  & 283  & 2.61  & 0.325  & $\mathbf{\underline{506}}$  & 1.21  & 0.334  & 497  & $\mathbf{\underline{0.74}}$  & 0.349 \\
0060 & 528 & $ 41.6\% $  & 518  & $\mathbf{\underline{0.10}}$  & $\mathbf{\underline{0.026}}$  & 503  & 0.29  & 0.029  & $\mathbf{\underline{525}}$  & 0.85  & 0.124  & 520  & 0.11  & 0.048 \\
0197 & 870 & $ 40.7\% $  & 718  & 0.58  & $\mathbf{\underline{0.108}}$  & 667  & 4.22  & 0.333  & $\mathbf{\underline{855}}$  & 1.16  & 0.227  & 813  & $\mathbf{\underline{0.43}}$  & 0.130 \\
0094 & 763 & $ 40.1\% $  & 659  & 0.81  & $\mathbf{\underline{0.116}}$  & 537  & 3.77  & 0.750  & $\mathbf{\underline{742}}$  & $\mathbf{\underline{0.75}}$  & 0.160  & 711  & 0.88  & 3.907 \\
0265 & 571 & $ 38.8\% $  & 372  & 5.30  & 1.433  & 346  & $\mathbf{\underline{1.25}}$  & $\mathbf{\underline{0.389}}$  & $\mathbf{\underline{554}}$  & 5.83  & 2.216  & $\mathbf{\underline{554}}$  & 7.46  & 2.839 \\
0083 & 635 & $ 31.3\% $  & 622  & 0.32  & 0.062  & 596  & 0.64  & 0.058  & $\mathbf{\underline{632}}$  & 0.37  & 0.372  & 614  & $\mathbf{\underline{0.08}}$  & $\mathbf{\underline{0.016}}$ \\
0076 & 558 & $ 30.5\% $  & 547  & $\mathbf{\underline{0.09}}$  & $\mathbf{\underline{0.037}}$  & 524  & 0.37  & 0.094  & $\mathbf{\underline{549}}$  & 0.78  & 0.120  & 540  & 0.17  & 0.042 \\
0185 & 368 & $ 30.0\% $  & 364  & 0.12  & 0.021  & 350  & $\mathbf{\underline{0.06}}$  & $\mathbf{\underline{0.010}}$  & $\mathbf{\underline{365}}$  & 0.41  & 0.094  & $\mathbf{\underline{365}}$  & 0.16  & 0.051 \\
0048 & 512 & $ 24.2\% $  & 501  & $\mathbf{\underline{0.10}}$  & $\mathbf{\underline{0.011}}$  & 474  & 4.69  & 0.178  & $\mathbf{\underline{507}}$  & 0.41  & 0.105  & 505  & 0.15  & 0.224 \\
0024 & 356 & $ 23.0\% $  & 328  & 3.59  & 1.122  & 309  & 2.03  & 0.398  & $\mathbf{\underline{355}}$  & 0.56  & 0.219  & 338  & $\mathbf{\underline{0.15}}$  & $\mathbf{\underline{0.104}}$ \\
0223 & 214 & $ 17.0\% $  & 211  & 3.45  & 0.285  & 204  & 3.76  & 0.510  & 212  & 3.34  & 0.519  & $\mathbf{\underline{213}}$  & $\mathbf{\underline{1.75}}$  & $\mathbf{\underline{0.275}}$ \\
5016 & 28 & $ 16.9\% $  & $\mathbf{\underline{28}}$  & $\mathbf{\underline{0.08}}$  & $\mathbf{\underline{0.015}}$  & $\mathbf{\underline{28}}$  & 0.12  & 0.016  & $\mathbf{\underline{28}}$  & 0.10  & 0.061  & $\mathbf{\underline{28}}$  & 0.08  & 0.046 \\
0046 & 440 & $ 14.6\% $  & 438  & 0.47  & 0.073  & 399  & 0.95  & 0.043  & 434  & 0.25  & 0.112  & $\mathbf{\underline{440}}$  & $\mathbf{\underline{0.03}}$  & $\mathbf{\underline{0.007}}$ \\
\midrule \midrule
0099 & 299 & $ 47.4\% $  & 157  & 0.56  & 0.229  & 190  & 3.53  & 0.709  & $\mathbf{\underline{297}}$  & 3.28  & 0.664  & 255  & $\mathbf{\underline{0.15}}$  & $\mathbf{\underline{0.085}}$ \\
1001 & 285 & $ 43.9\% $  & 268  & 3.87  & 2.585  & 251  & $\mathbf{\underline{1.70}}$  & $\mathbf{\underline{0.661}}$  & $\mathbf{\underline{275}}$  & 7.89  & 3.990  & 270  & 4.56  & 3.817 \\
0231 & 296 & $ 42.2\% $  & 260  & $\mathbf{\underline{0.31}}$  & $\mathbf{\underline{0.029}}$  & 246  & 0.84  & 0.065  & $\mathbf{\underline{286}}$  & 1.37  & 0.322  & 278  & 0.73  & 0.134 \\
0411 & 299 & $ 29.9\% $  & 268  & 0.23  & 0.036  & 273  & $\mathbf{\underline{0.13}}$  & $\mathbf{\underline{0.020}}$  & $\mathbf{\underline{293}}$  & 0.39  & 0.196  & 268  & 0.19  & 0.148 \\
0377 & 295 & $ 27.5\% $  & 232  & $\mathbf{\underline{0.12}}$  & $\mathbf{\underline{0.016}}$  & 210  & 0.29  & 0.018  & $\mathbf{\underline{269}}$  & 1.13  & 0.205  & 266  & 0.65  & 0.237 \\
0102 & 299 & $ 25.8\% $  & $\mathbf{\underline{296}}$  & 0.18  & $\mathbf{\underline{0.023}}$  & 284  & 0.28  & 0.059  & 294  & 2.31  & 0.698  & 293  & $\mathbf{\underline{0.15}}$  & 0.101 \\
0147 & 298 & $ 24.6\% $  & $\mathbf{\underline{289}}$  & $\mathbf{\underline{1.36}}$  & $\mathbf{\underline{0.118}}$  & 207  & 4.62  & 0.325  & 284  & 6.36  & 0.934  & $\mathbf{\underline{289}}$  & 6.75  & 3.542 \\
0148 & 287 & $ 24.6\% $  & 244  & 1.15  & 0.135  & 197  & $\mathbf{\underline{0.60}}$  & $\mathbf{\underline{0.035}}$  & 275  & 13.98  & 1.558  & $\mathbf{\underline{282}}$  & 22.73  & 2.646 \\
0446 & 298 & $ 22.1\% $  & $\mathbf{\underline{296}}$  & 0.34  & $\mathbf{\underline{0.023}}$  & 288  & 0.71  & 0.046  & 289  & 1.23  & 0.391  & 295  & $\mathbf{\underline{0.20}}$  & 0.071 \\
0022 & 297 & $ 21.2\% $  & 280  & $\mathbf{\underline{0.11}}$  & $\mathbf{\underline{0.015}}$  & 274  & 0.29  & 0.039  & $\mathbf{\underline{296}}$  & 0.58  & 0.160  & 280  & 0.22  & 0.087 \\
0327 & 298 & $ 21.0\% $  & 280  & 0.48  & $\mathbf{\underline{0.032}}$  & 271  & $\mathbf{\underline{0.26}}$  & 0.090  & $\mathbf{\underline{288}}$  & 1.27  & 0.360  & $\mathbf{\underline{288}}$  & 15.54  & 2.035 \\
0015 & 284 & $ 20.6\% $  & 241  & 0.61  & $\mathbf{\underline{0.083}}$  & 215  & 1.04  & 0.167  & 244  & 2.21  & 0.389  & $\mathbf{\underline{272}}$  & $\mathbf{\underline{0.28}}$  & 0.095 \\
0455 & 298 & $ 19.8\% $  & 292  & 0.40  & 0.071  & 293  & 0.68  & 0.105  & 294  & 0.77  & 0.159  & $\mathbf{\underline{297}}$  & $\mathbf{\underline{0.35}}$  & $\mathbf{\underline{0.064}}$ \\
0496 & 297 & $ 19.2\% $  & 280  & 0.69  & $\mathbf{\underline{0.041}}$  & 281  & $\mathbf{\underline{0.35}}$  & 0.055  & 285  & 1.40  & 0.550  & $\mathbf{\underline{291}}$  & 0.44  & 0.303 \\
1589 & 299 & $ 17.4\% $  & 294  & 0.13  & 0.074  & 290  & 0.14  & $\mathbf{\underline{0.019}}$  & 288  & 0.82  & 0.193  & $\mathbf{\underline{298}}$  & $\mathbf{\underline{0.07}}$  & 0.041 \\
0012 & 299 & $ 16.3\% $  & $\mathbf{\underline{295}}$  & 0.77  & 0.086  & 287  & $\mathbf{\underline{0.40}}$  & $\mathbf{\underline{0.027}}$  & 129  & 1.04  & 0.318  & $\mathbf{\underline{295}}$  & 0.51  & 0.121 \\
0104 & 284 & $ 16.2\% $  & 237  & 4.04  & 0.330  & 193  & $\mathbf{\underline{0.29}}$  & $\mathbf{\underline{0.029}}$  & 265  & 17.05  & 1.530  & $\mathbf{\underline{280}}$  & 19.69  & 0.834 \\
0019 & 299 & $ 15.4\% $  & 288  & 0.23  & $\mathbf{\underline{0.008}}$  & 250  & $\mathbf{\underline{0.06}}$  & $\mathbf{\underline{0.008}}$  & 271  & 0.81  & 0.250  & $\mathbf{\underline{296}}$  & 0.09  & 0.025 \\
0063 & 293 & $ 14.5\% $  & 266  & $\mathbf{\underline{0.14}}$  & $\mathbf{\underline{0.024}}$  & 262  & 0.46  & 0.048  & 268  & 0.92  & 0.605  & $\mathbf{\underline{287}}$  & 0.32  & 0.100 \\
0130 & 285 & $ 14.4\% $  & 202  & 0.21  & $\mathbf{\underline{0.019}}$  & 192  & $\mathbf{\underline{0.20}}$  & 0.023  & 187  & 1.20  & 0.349  & $\mathbf{\underline{279}}$  & 2.00  & 0.909 \\
0080 & 284 & $ 12.9\% $  & 141  & $\mathbf{\underline{0.10}}$  & $\mathbf{\underline{0.017}}$  & 139  & 0.59  & 0.096  & 278  & 2.62  & 0.868  & $\mathbf{\underline{282}}$  & 1.92  & 0.236 \\
0240 & 298 & $ 11.9\% $  & 288  & 0.64  & $\mathbf{\underline{0.088}}$  & 275  & 3.13  & 0.265  & 285  & 1.31  & 0.470  & $\mathbf{\underline{290}}$  & $\mathbf{\underline{0.39}}$  & 0.135 \\
0007 & 290 & $ 11.7\% $  & 284  & 1.51  & 0.079  & 172  & 0.91  & 0.041  & 277  & 1.24  & 0.174  & $\mathbf{\underline{289}}$  & $\mathbf{\underline{0.19}}$  & $\mathbf{\underline{0.035}}$ \\

    \end{tabular}
    \end{adjustbox}

\end{table*}

\subsection{Results}
\label{subsec:results}

Our results on the MegaDepth and 1DSfM test sets and comparisons to baselines are shown in Tables~\ref{tab:megadepth_mean_errors} and~\ref{tab:1dsfm_mean_errors}, respectively. 
For each scene, we also report the number of input images ($N_c$), the fraction of outlier track points, and compare our \method{} method against the baselines in terms of number of registered images, mean rotation error (in degrees), translation error, and runtime.

Across both benchmarks, \method{} outperforms the deep factorization baseline \resfm{} on most scenes, achieving lower rotation and translation errors. Compared to classical pipelines, \method{} is competitive with \theia{} and \glomap{}, and often surpasses them on both metrics. In terms of coverage, \method{} registers a larger fraction of images than \resfm{}, though typically fewer than \glomap{}.

\begin{table*}[]

\caption{{\small {\bf 1DSfM experiment.} For each scene, we show the number of input images (denoted $N_c$) and the fraction of outliers. For each model, we show the number of images used for reconstruction (denoted $N_r$) and mean values of the rotation error (in degrees) and translation error. Winning results are marked in \textbf{\underline{bold and underlined}}.}}

    \label{tab:1dsfm_mean_errors}
    \centering
    \scriptsize 
    \rowcolors{3}{rowgray}{white} 
    \setlength{\tabcolsep}{3pt} 
    \renewcommand{\arraystretch}{1.5} 
    \begin{adjustbox}{max width=\textwidth}
    \begin{tabular}{lcc|ccc|ccc||ccc|ccc}
    
\multirow{2}{*}{ Scene } &
\multirow{2}{*}{ $N_c$ } &
\multirow{2}{*}{ Outliers\% } &
\multicolumn{3}{c}{\textbf{Ours}}&
\multicolumn{3}{c}{RESFM} &
\multicolumn{3}{c}{\theia}&
\multicolumn{3}{c}{\glomap}\\

 &  &  & $N_r$ & Rot  & Trans   & $N_r$ & Rot  & Trans   &$N_r$ & Rot  & Trans    & $N_r$ & Rot  & Trans \\
Alamo & 573 & $ 32.6\% $  & 523  & $\mathbf{\underline{1.35}}$  & $\mathbf{\underline{0.322}}$  & 484  & 3.66  & 0.515  & 549  & 4.42  & 1.433  & $\mathbf{\underline{557}}$  & 2.45  & 1.520 \\
Ellis Island & 227 & $ 25.1\% $  & 215  & $\mathbf{\underline{0.28}}$  & $\mathbf{\underline{0.081}}$  & 214  & 0.82  & 0.122  & 213  & 5.01  & 1.527  & $\mathbf{\underline{219}}$  & 0.58  & 0.155 \\
Madrid Metropolis & 333 & $ 39.4\% $  & 298  & 1.31  & $\mathbf{\underline{0.145}}$  & 244  & 8.42  & 0.827  & 319  & 2.61  & 0.903  & $\mathbf{\underline{320}}$  & $\mathbf{\underline{1.22}}$  & 0.242 \\
Montreal Notre Dame & 448 & $ 31.7\% $  & 442  & 0.93  & 0.418  & 346  & 2.82  & 0.352  & 420  & 4.47  & 1.285  & $\mathbf{\underline{444}}$  & $\mathbf{\underline{0.60}}$  & $\mathbf{\underline{0.211}}$ \\
NYC Library & 330 & $ 33.6\% $  & 295  & $\mathbf{\underline{0.34}}$  & $\mathbf{\underline{0.095}}$  & 224  & 3.96  & 0.429  & 313  & 4.06  & 1.141  & $\mathbf{\underline{323}}$  & 0.58  & 0.189 \\
Notre Dame & 549 & $ 35.6\% $  & 527  & $\mathbf{\underline{0.52}}$  & $\mathbf{\underline{0.105}}$  & 517  & 1.20  & 0.231  & 531  & 3.70  & 0.828  & $\mathbf{\underline{543}}$  & 2.73  & 0.389 \\
Piazza del Popolo & 336 & $ 33.1\% $  & 318  & 5.37  & 0.670  & 249  & 2.20  & $\mathbf{\underline{0.186}}$  & 324  & 3.31  & 1.053  & $\mathbf{\underline{331}}$  & $\mathbf{\underline{0.80}}$  & 0.188 \\
Tower of London & 467 & $ 27.0\% $  & 457  & 0.89  & 0.057  & 94  & $\mathbf{\underline{0.67}}$  & $\mathbf{\underline{0.026}}$  & 448  & 6.61  & 1.189  & $\mathbf{\underline{466}}$  & 0.81  & 0.138 \\
Vienna Cathedral & 824 & $ 31.4\% $  & 763  & 7.90  & 0.590  & 479  & $\mathbf{\underline{1.52}}$  & $\mathbf{\underline{0.112}}$  & 767  & 12.25  & 1.663  & $\mathbf{\underline{822}}$  & 2.00  & 2.414 \\
Yorkminster & 432 & $ 29.0\% $  & 402  & $\mathbf{\underline{0.43}}$  & $\mathbf{\underline{0.044}}$  & 331  & 14.54  & 1.468  & 387  & 8.35  & 1.916  & $\mathbf{\underline{418}}$  & 0.95  & 0.316 \\

    \end{tabular}
    \end{adjustbox}

\end{table*}

For completeness, we also report the performance of several recent deep learning-based pose estimation methods that aim to scale to larger image collections than those supported by approaches such as \vggt{}. The per-scene results are shown in Table~\ref{tabapp:1dsfm_deep}. While these methods are computationally efficient, their pose accuracy on this benchmark remains noticeably lower than that of SfM-based pipelines.

\begin{table}[]
\caption{{\small {\bf Deep-based methods on the 1DSfM dataset.} 
For each scene we list the number of input images ($N_c$). 
For each deep model (TTT3R, CUT3R, FAST3R) we report the mean rotation error (degrees) and mean translation error. 
Best results are \textbf{\underline{bold and underlined}}.}}

\label{tabapp:1dsfm_deep}
\centering
\tiny
\rowcolors{3}{rowgray}{white} 
\setlength{\tabcolsep}{5pt}
\renewcommand{\arraystretch}{1.0}
\begin{tabular}{l c | cc | cc | cc}
\toprule
\multirow{2}{*}{Scene} & 
\multirow{2}{*}{$N_c$} &
\multicolumn{2}{c}{TTT3R} &
\multicolumn{2}{c}{CUT3R} &
\multicolumn{2}{c}{FAST3R} \\
& & Rot & Trans & Rot & Trans & Rot & Trans \\
\midrule

Alamo & 573 &
\underline{\textbf{16.08}} & \underline{\textbf{3.650}} &
22.32 & 4.266 &
40.37 & 3.990 \\

Ellis Island & 227 &
\underline{\textbf{8.85}} & \underline{\textbf{2.333}} &
12.53 & 3.171 &
11.74 & 3.201 \\

Madrid Metropolis & 333 &
\underline{\textbf{14.83}} & \underline{\textbf{2.258}} &
18.81 & 3.189 &
67.37 & 3.574 \\

Montreal Notre Dame & 448 &
13.25 & \underline{\textbf{1.230}} &
16.12 & 2.134 &
\underline{\textbf{10.79}} & 2.052 \\

NYC Library & 330 &
\underline{\textbf{7.67}} & \underline{\textbf{1.656}} &
10.26 & 1.912 &
11.90 & 2.408 \\

Notre Dame & 549 &
\underline{\textbf{11.88}} & \underline{\textbf{1.430}} &
15.33 & 1.694 &
16.44 & 2.241 \\

Piazza del Popolo & 336 &
23.13 & \underline{\textbf{2.063}} &
\underline{\textbf{22.63}} & 2.092 &
32.48 & 2.568 \\

Tower of London & 467 &
\underline{\textbf{29.69}} & 3.647 &
29.85 & \underline{\textbf{3.607}} &
59.53 & 3.658 \\

Vienna Cathedral & 824 &
43.76 & 2.978 &
41.58 & 2.802 &
\underline{\textbf{29.49}} & \underline{\textbf{2.561}} \\

Yorkminster & 432 &
\underline{\textbf{16.45}} & \underline{\textbf{2.223}} &
23.00 & 2.992 &
20.26 & 2.463 \\

\bottomrule
\end{tabular}
\end{table}

Following \cite{khatib2025resfm}, we evaluate \method{} on the smaller Strecha and BlendedMVS benchmarks, which provide ground-truth camera poses. As shown in Table~\ref{tab:strecha_blendedmvs}, \method{} is consistently more accurate than image-based deep baselines (\vggsfm{}, MASt3R, and \vggt{}), which typically do not scale to the larger datasets considered, and it performs on par with classical pipelines (including \theia{}, \colmap{}, and \glomap{}).

\begin{table*}[h!] \captionsetup{font=normalsize} \caption{{\small \textbf{Strecha \& BlendedMVS datasets.} For each scene we list the number of input images ($N_c$) and outlier fraction. For each method we report the number of registered images ($N_r$), mean rotation error (deg), translation error, and runtime (s). Best is \textbf{bold}, second best is \underline{underlined}.}} \label{tab:strecha_blendedmvs} \centering \large \rowcolors{3}{rowgray}{white} \begin{adjustbox}{max width=\textwidth, max totalheight=\textheight} \setlength{\tabcolsep}{1pt} \fontsize{10pt}{12pt}\selectfont \begin{tabular}{lcc|cccc|cccc|cccc|cccc|cccc|cccc|cccc} \multirow{2}{*}{Scene} & \multirow{2}{*}{$N_c$} & \multirow{2}{*}{Out.\%} & \multicolumn{4}{c}{\textbf{Ours}} & \multicolumn{4}{c}{VGGT} & \multicolumn{4}{c}{MASt3R} & \multicolumn{4}{c}{VGGSfM} & \multicolumn{4}{c}{\theia} & \multicolumn{4}{c}{\colmap} & \multicolumn{4}{c}{\glomap} \\ ~ & ~ & ~ & $N_r$ & Rot & Trans & Time & $N_r$ & Rot & Trans & Time & $N_r$ & Rot & Trans & Time & $N_r$ & Rot & Trans & Time & $N_r$ & Rot & Trans & Time & $N_r$ & Rot & Trans & Time & $N_r$ & Rot & Trans & Time \\ \hline \multicolumn{31}{c}{\textbf{Strecha}} \\ entry-P10 & 10 & 4.8 & 10 & \textbf{0.004} & \textbf{0.0005} & \underline{7} & 10 & 0.079 & 0.033 & 16.5 & 10 & 0.442 & 0.055 & 19 & 10 & 0.165 & 0.056 & 10.3 & 10 & 0.024 & 0.008 & \textbf{0.9} & 10 & \underline{0.023} & \underline{0.007} & 36.0 & 10 & 0.187 & 0.026 & 12.5 \\ fountain-P11 & 11 & 1.4 & 11 & \textbf{0.009} & \textbf{0.0003} & \underline{8} & 11 & 0.034 & 0.019 & 12.2 & 11 & 0.160 & 0.026 & 22 & 11 & 0.172 & 0.016 & 15.4 & 11 & \underline{0.027} & \underline{0.002} & \textbf{1.5} & 11 & \underline{0.027} & 0.003 & 37.0 & 11 & 0.194 & 0.022 & 38.6 \\ Herz-Jesus-P8 & 8 & 1.8 & 8 & \textbf{0.009} & \textbf{0.0010} & 6 & 8 & 0.032 & 0.011 & 12.7 & 8 & 0.363 & 0.037 & 16 & 8 & 0.206 & 0.042 & 8.7 & 8 & \underline{0.025} & 0.005 & \textbf{0.6} & 8 & 0.026 & \underline{0.004} & 22.0 & 8 & 0.091 & 0.015 & \underline{5.0} \\ Herz-Jesus-P25 & 25 & 2.8 & 25 & \textbf{0.010} & \textbf{0.0003} & \underline{12} & 25 & 0.048 & 0.007 & 31.9 & 25 & 0.869 & 0.057 & 81 & 25 & 0.158 & 0.046 & 19.6 & 25 & \underline{0.026} & \underline{0.006} & \textbf{2.4} & 25 & 0.028 & \underline{0.006} & 60.0 & 25 & 0.138 & 0.013 & 76.6 \\ \hline \multicolumn{31}{c}{\textbf{BlendedMVS}} \\ scene0 & 75 & 2.0 & 74 & 0.149 & 0.0204 & 73 & 75 & 0.041 & 0.017 & 108 & 75 & 0.501 & 0.191 & 516 & 75 & 0.045 & 0.0106 & \underline{61} & 75 & 0.009 & 0.0017 & \textbf{49} & 75 & \textbf{0.006} & \textbf{0.0005} & 106 & 75 & \underline{0.007} & \underline{0.0016} & 198 \\ scene1 & 51 & 1.4 & 51 & 0.342 & 0.0345 & \underline{28} & 51 & 0.101 & 0.050 & 41 & 51 & 0.919 & 0.173 & 1017 & 51 & 0.098 & 0.0112 & 32 & 51 & 0.029 & \underline{0.0099} & \textbf{18} & 51 & \textbf{0.007} & \textbf{0.0003} & 67 & 51 & \underline{0.024} & 0.0102 & 117 \\ scene2 & 33 & 2.2 & 33 & \underline{0.008} & \underline{0.0004} & \underline{17} & 33 & 0.230 & 0.022 & 52 & 33 & 1.972 & 0.130 & 117 & 33 & 0.227 & 0.0180 & 30 & 33 & 0.045 & 0.0098 & \textbf{15} & 33 & \textbf{0.003} & \textbf{0.0002} & 55 & 33 & 0.025 & 0.0060 & 87 \\ scene3 & 66 & 8.8 & 66 & \underline{0.006} & \underline{0.0003} & 54 & 66 & 0.353 & 0.014 & 276 & 66 & 0.927 & 0.045 & 815 & 66 & 0.372 & 0.0174 & \underline{52} & 66 & 0.019 & 0.0018 & \textbf{21} & 66 & \textbf{0.004} & \textbf{0.0002} & 128 & 66 & 0.008 & 0.0017 & 392 \\\end{tabular} \end{adjustbox} \end{table*}

\noindent\textbf{Robustness to retrieval-based graph construction.}
We train \method{} using relative poses obtained from \textbf{exhaustive} pairwise matching. At test time, we additionally evaluate a sparse retrieval-based view graph, where each image is connected to its top-30 neighbors retrieved by MegaLoc~\cite{berton2025megaloc}. As shown in Table~\ref{tab:graph_density}, \method{} often preserves strong registration coverage and accuracy despite the large reduction in edge density.

\begin{table*}[t]
\caption{{\small \textbf{Robustness to retrieval-based graph construction.}
We evaluate VGPA on the full 1DSfM benchmark using either exhaustive pairwise matching or a sparse retrieval-based view graph constructed with MegaLoc top-30 retrieval.
For each scene, we report the number of input images ($N_c$), the number of registered images ($N_r$), mean rotation error (deg), and mean translation error. Best results are marked in \textbf{\underline{bold and underlined}}.}}
\label{tab:graph_density}
\centering
\scriptsize
\rowcolors{4}{rowgray}{white}
\setlength{\tabcolsep}{3pt}
\renewcommand{\arraystretch}{1.2}
\begin{adjustbox}{max width=\textwidth}
\begin{tabular}{l c | c c c | c c c}

\multirow{3}{*}{Scene} &
\multirow{3}{*}{$N_c$} &
\multicolumn{6}{c}{\textbf{Ours}} \\

& & \multicolumn{3}{c|}{\textbf{Exhaustive Matching}} &
\multicolumn{3}{c}{\textbf{MegaLoc@30}} \\

& & $N_r$ & Rot & Trans &
$N_r$ & Rot & Trans \\
\midrule

Alamo & 573 & 523 & 1.35 & 0.322 & $\mathbf{\underline{548}}$ & $\mathbf{\underline{1.18}}$ & $\mathbf{\underline{0.163}}$ \\
Ellis Island & 227 & 215 & 0.28 & $\mathbf{\underline{0.081}}$ & $\mathbf{\underline{216}}$ & $\mathbf{\underline{0.21}}$ & 0.090 \\
Madrid Metropolis & 333 & 298 & $\mathbf{\underline{1.31}}$ & $\mathbf{\underline{0.145}}$ & $\mathbf{\underline{323}}$ & 1.48 & 3.397 \\
Montreal Notre Dame & 448 & $\mathbf{\underline{442}}$ & 0.93 & 0.418 & 338 & $\mathbf{\underline{0.11}}$ & $\mathbf{\underline{0.015}}$ \\
NYC Library & 330 & 295 & $\mathbf{\underline{0.34}}$ & $\mathbf{\underline{0.095}}$ & $\mathbf{\underline{319}}$ & 0.92 & 0.537 \\
Notre Dame & 549 & 527 & 0.52 & 0.105 & $\mathbf{\underline{535}}$ & $\mathbf{\underline{0.42}}$ & $\mathbf{\underline{0.086}}$ \\
Piazza del Popolo & 336 & 318 & 5.37 & 0.670 & $\mathbf{\underline{329}}$ & $\mathbf{\underline{0.57}}$ & $\mathbf{\underline{0.064}}$ \\
Tower of London & 467 & $\mathbf{\underline{457}}$ & $\mathbf{\underline{0.89}}$ & $\mathbf{\underline{0.057}}$ & 447 & 2.14 & 0.408 \\
Vienna Cathedral & 824 & 763 & $\mathbf{\underline{7.90}}$ & 0.590 & $\mathbf{\underline{771}}$ & 9.32 & $\mathbf{\underline{0.462}}$ \\
Yorkminster & 432 & 402 & $\mathbf{\underline{0.43}}$ & $\mathbf{\underline{0.044}}$ & $\mathbf{\underline{405}}$ & 6.32 & 0.469 \\

\end{tabular}
\end{adjustbox}
\end{table*}

\noindent\textbf{Uncalibrated image collections.}
Table~\ref{tab:uncalibrated} compares two settings: (i) using ground-truth intrinsics and (ii) starting from an approximate calibration ($f_x{,}f_y$ proportional to image size, principal point at the image center) and optimizing intrinsics jointly with the extrinsics during bundle adjustment. More details on this protocol are provided in the supplementary material. While self-calibration incurs a small accuracy drop relative to ground-truth intrinsics, \method{} remains competitive and maintains high performance.

\begin{table*}[]
      \caption{{\small \textbf{Impact of Camera Intrinsics (Known vs.\ Estimated).}
      For each scene, we report the number of input images ($N_c$) and the outlier fraction.
      We compare our method with known intrinsics vs.\ without intrinsics and report
      $N_r$, mean rotation error (deg), and mean translation error. Best results are in \textbf{bold}.}}
      \label{tab:uncalibrated}
      \centering
      \tiny
      \rowcolors{3}{rowgray}{white}
      \begin{adjustbox}{max width=\textwidth}
      \begin{tabular}{lcc|ccc|ccc}
  \multirow{2}{*}{Scene} &
  \multirow{2}{*}{$N_c$} &
  \multirow{2}{*}{Out.\%} &
  \multicolumn{3}{c|}{\textbf{Ours (w/ intrinsics)}} &
  \multicolumn{3}{c}{\textbf{Ours (w/o intrinsics)}} \\
  ~ & ~ & ~ & $N_r$ & Rot & Trans & $N_r$ & Rot & Trans \\
  \midrule
  \multicolumn{9}{l}{\textit{BlendedMVS scenes (shared intrinsics)}} \\
  scene0 & 75 & 2.0 & \textbf{74} & 0.149 & 0.0204 & \textbf{74} & \textbf{0.122} & \textbf{0.0159} \\
  scene1 & 51 & 1.4 & \textbf{51} & 0.342 & \textbf{0.0345} & \textbf{51} & \textbf{0.337} & 0.0401 \\
  scene2 & 33 & 2.2 & \textbf{33} & \textbf{0.008} & \textbf{0.0004} & \textbf{33} & 0.025 & 0.0050 \\
  scene3 & 66 & 8.8 & \textbf{66} & \textbf{0.006} & \textbf{0.0003} & \textbf{66} & 0.010 & 0.0014 \\
  \midrule
  \multicolumn{9}{l}{\textit{MegaDepth scenes (not shared intrinsics)}} \\
  0012 & 299 & 16.3 & \textbf{295} & \textbf{0.77} & \textbf{0.086} & 292 & 0.98 & 0.253 \\
  0024 & 356 & 23.0 & \textbf{328} & 3.59 & 1.122 & 308 & \textbf{1.79} & \textbf{0.508} \\
  0048 & 512 & 24.2 & \textbf{501} & \textbf{0.10} & \textbf{0.011} & 485 & 0.38 & 0.456 \\
  0083 & 635 & 31.3 & \textbf{622} & \textbf{0.32} & 0.062 & 596 & 0.64 & \textbf{0.058} \\
      \end{tabular}
      \end{adjustbox}
  \end{table*}

\noindent\textbf{View Re-integration (optional post-processing).}
The results reported in the main paper are obtained \emph{without} this step. 
As an optional post-processing stage, we attempt to re-register views that may be discarded during the BA stage using a lightweight add-back procedure. 
Unregistered views are ranked by connectivity (e.g., number of 2D--3D matches) with the current point cloud. 
For each candidate, we estimate its pose from the available 2D--3D correspondences and refine it with a short local BA applied to its neighboring views. 
The process repeats until no further views can be added. 
Additional results comparing \emph{Ours} and \emph{Ours + post-processing} in terms of the number of registered images ($N_r$), mean rotation error (deg), and mean translation error are reported in the Supplementary Material. 
These results show that the add-back step increases the number of registered cameras with minimal runtime overhead (about 0.6 second per added view, without additional optimization).

\begin{figure}[]
    \centering
    \includegraphics[height=0.18\textwidth]{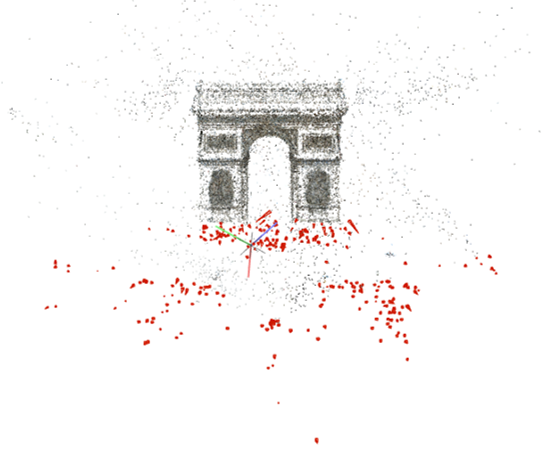}
    \includegraphics[height=0.18\textwidth]{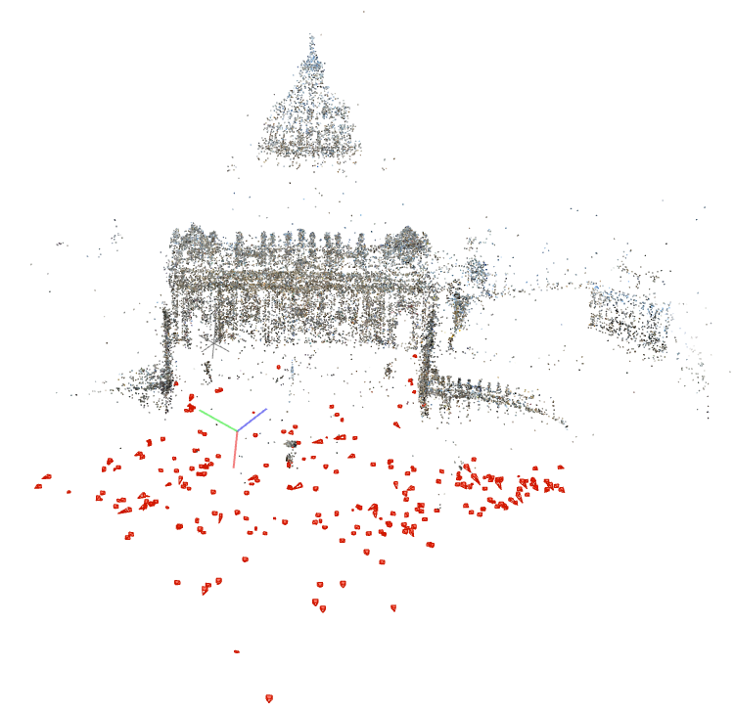}
    \includegraphics[height=0.18\textwidth]{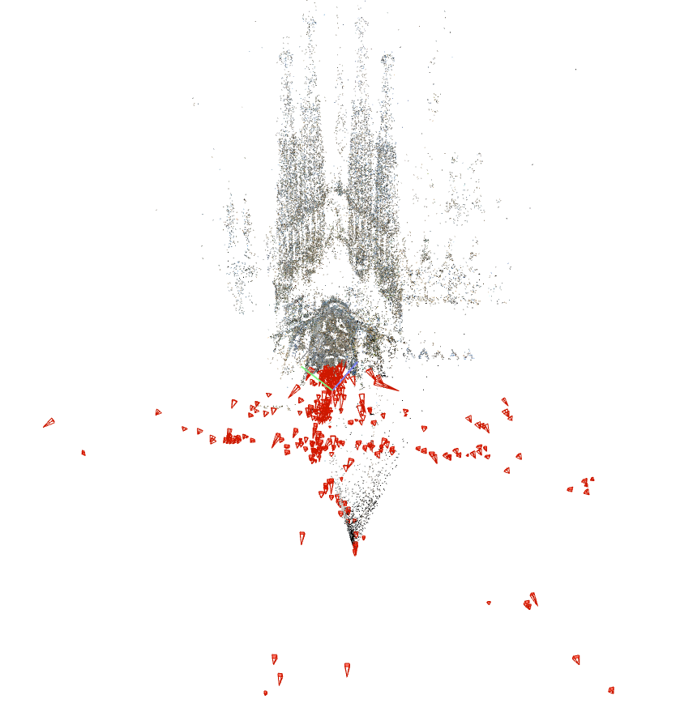}
    \includegraphics[height=0.18\textwidth]{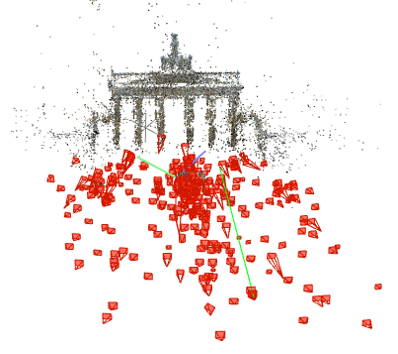}
    \includegraphics[height=0.18\textwidth]{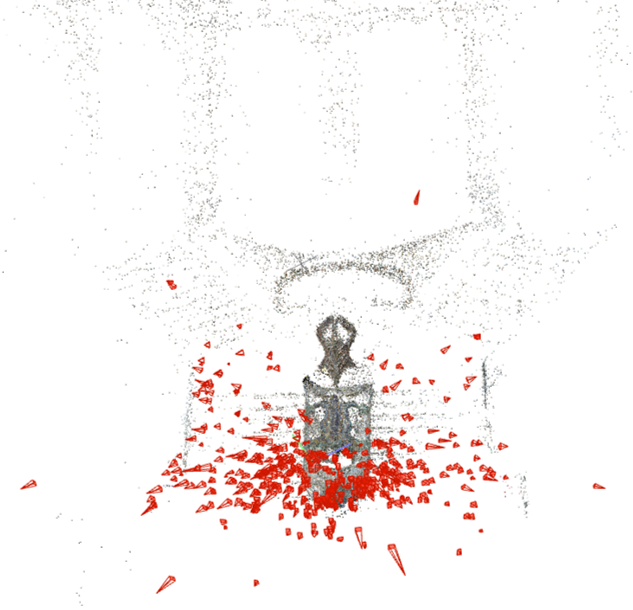}
    \includegraphics[height=0.18\textwidth]{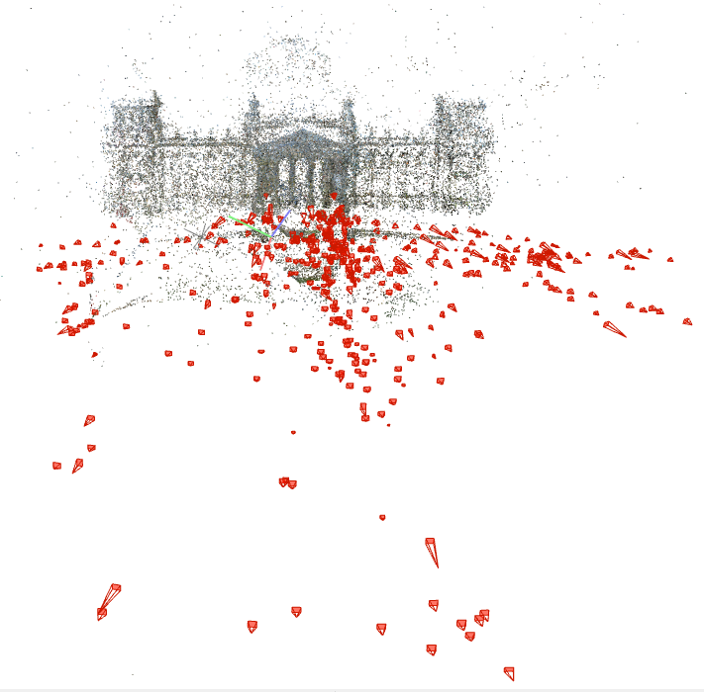}
    \includegraphics[height=0.18\textwidth]{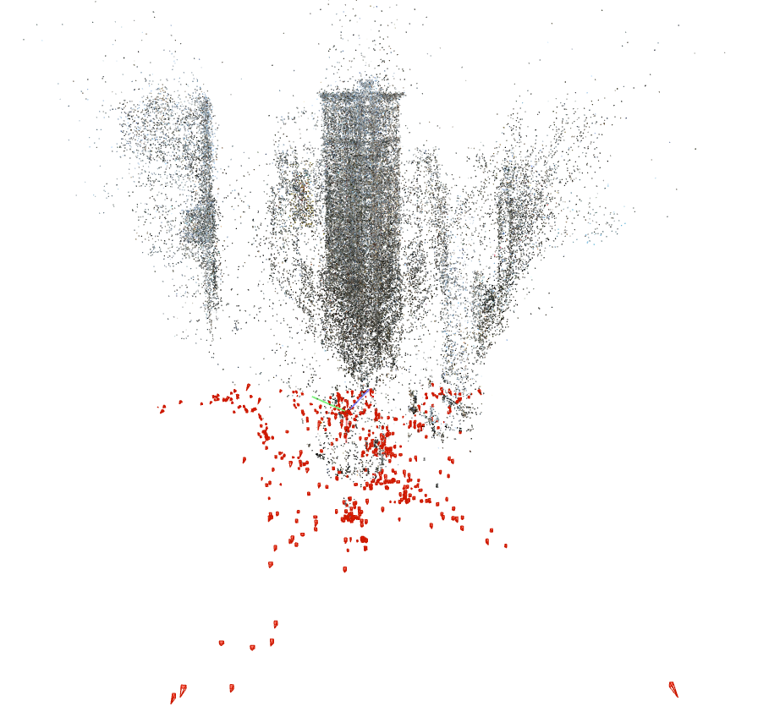}
    \includegraphics[height=0.18\textwidth]{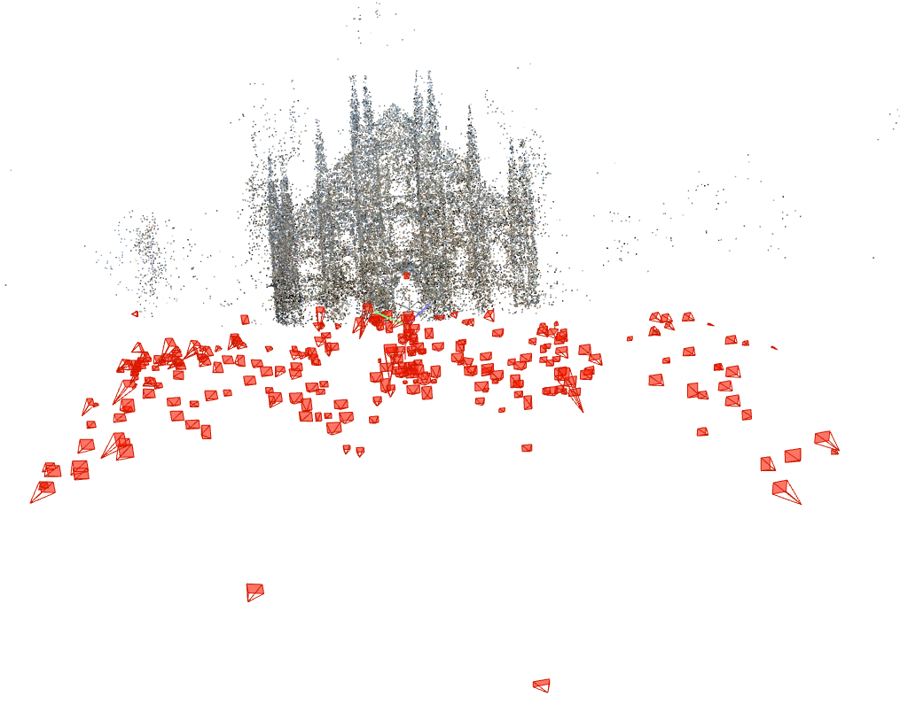}
    \includegraphics[height=0.18\textwidth]{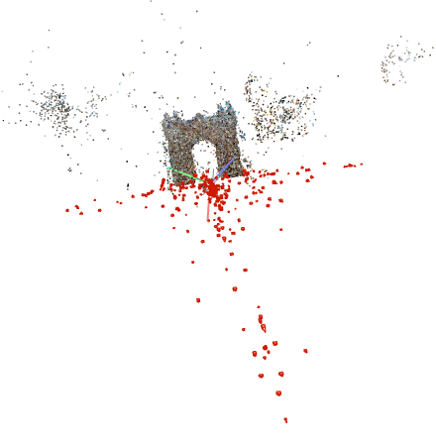}
    \includegraphics[height=0.18\textwidth]{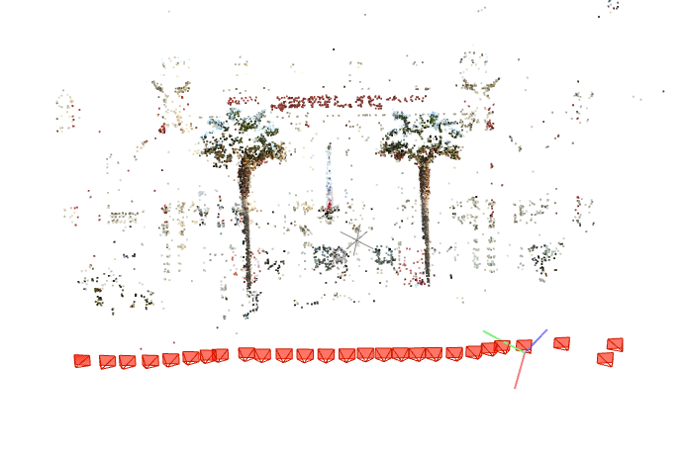}
    \includegraphics[height=0.18\textwidth]{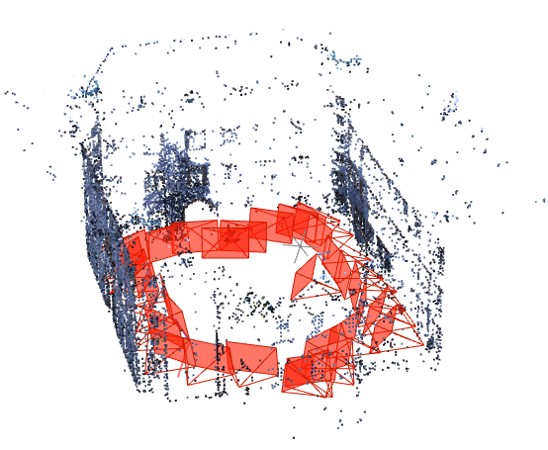}
    \includegraphics[height=0.18\textwidth]{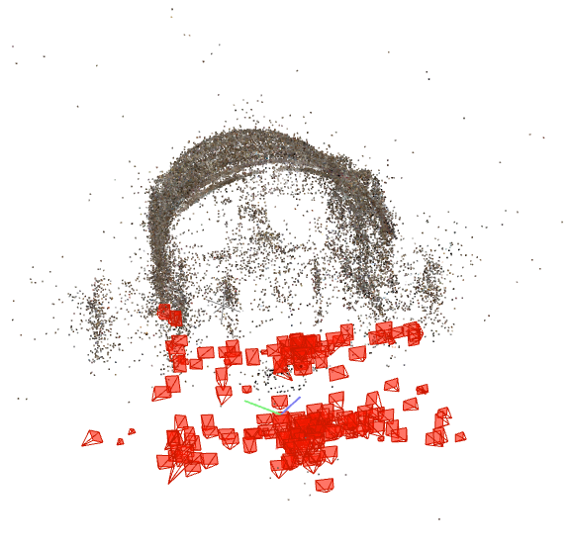}
    \includegraphics[height=0.18\textwidth]{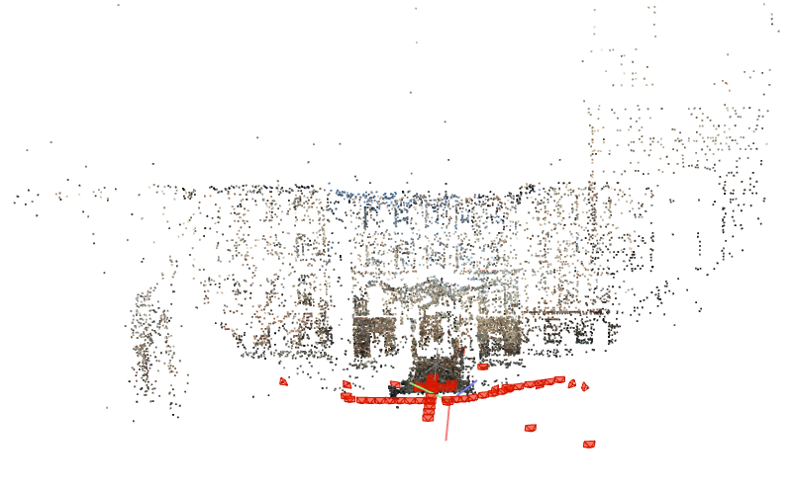}
    \includegraphics[height=0.18\textwidth]{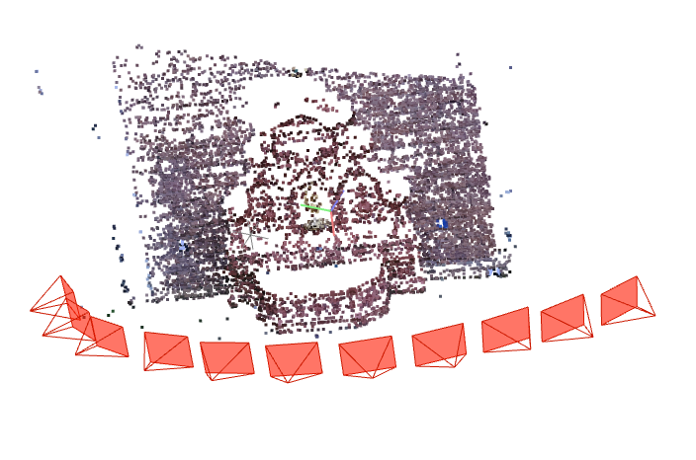}
    \includegraphics[height=0.18\textwidth]{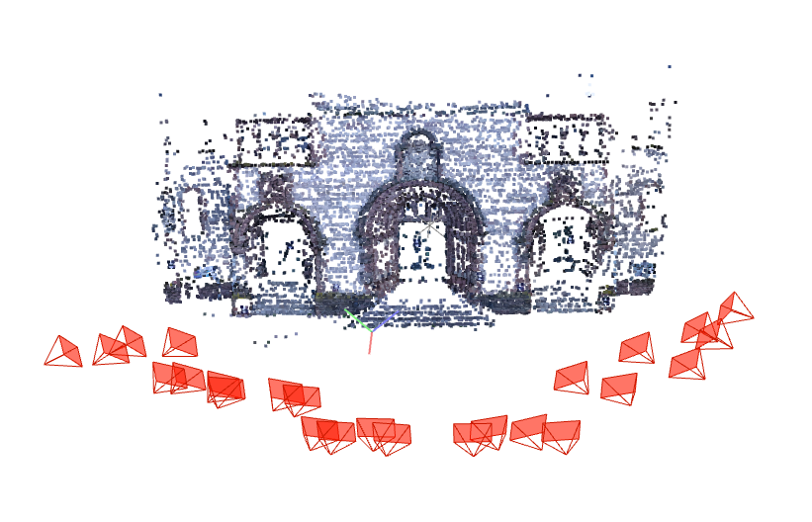}
    \includegraphics[height=0.18\textwidth]{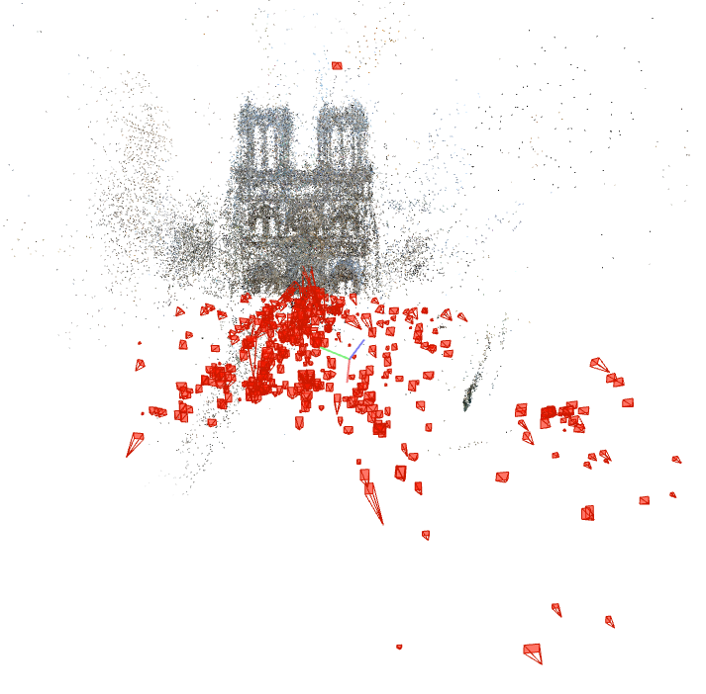}
    \includegraphics[height=0.18\textwidth]{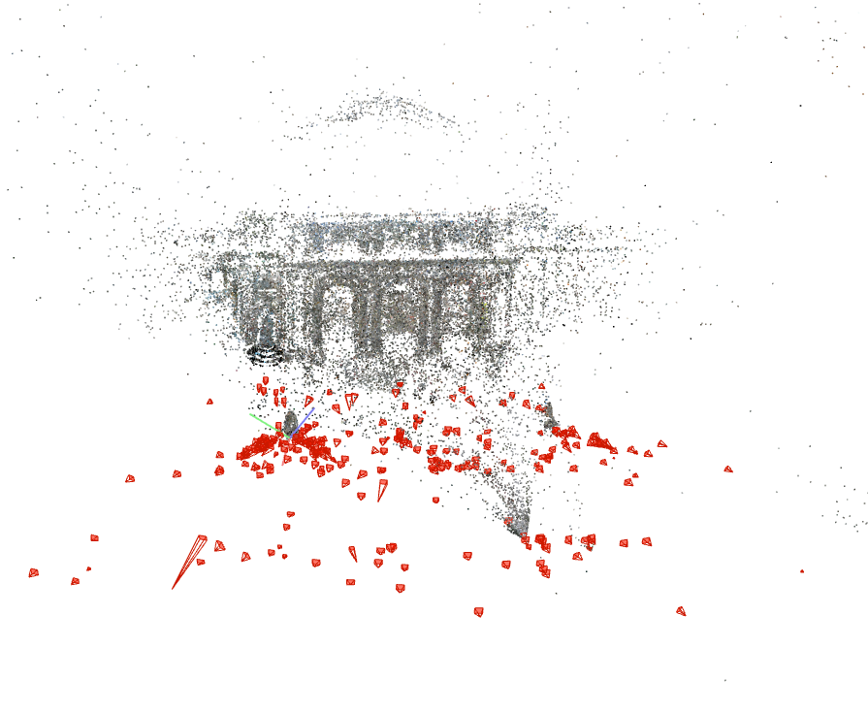}
    \includegraphics[height=0.18\textwidth]{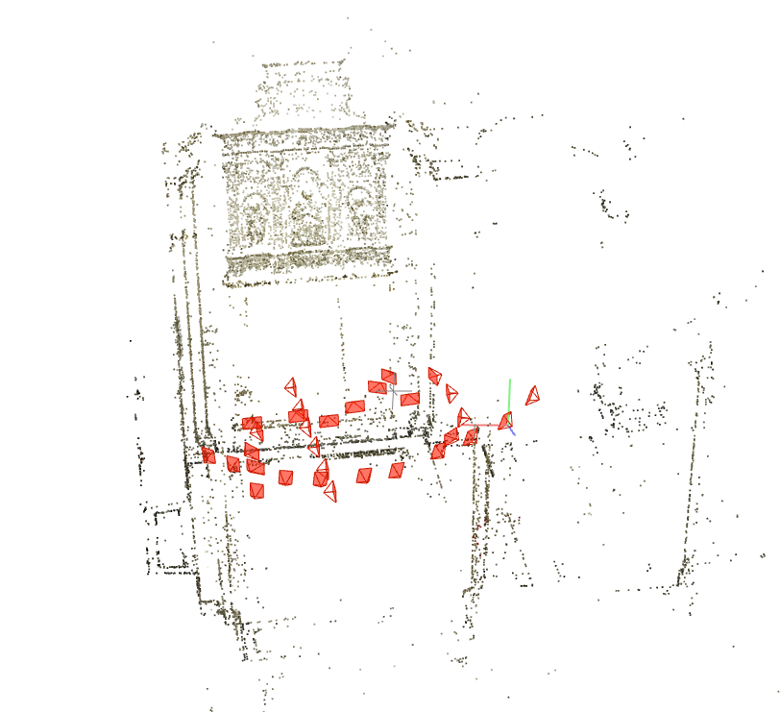}
    \includegraphics[height=0.18\textwidth]{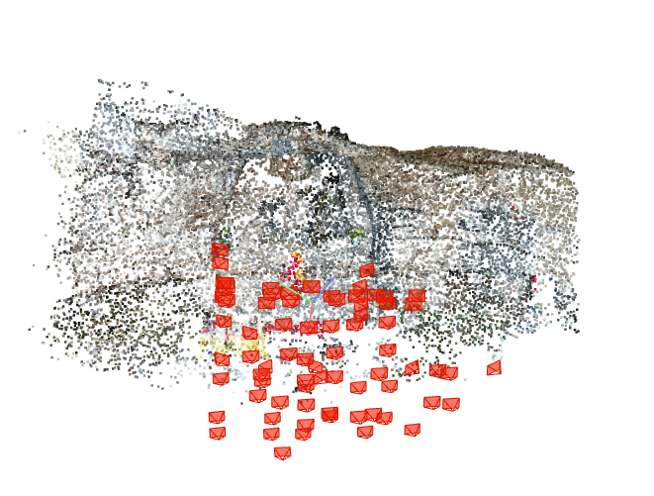}
    \includegraphics[height=0.18\textwidth]{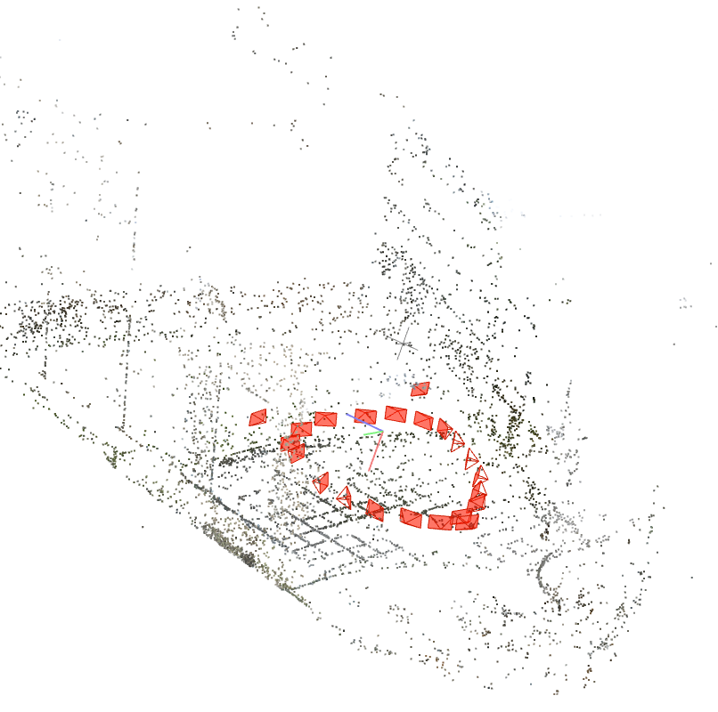}
    \includegraphics[height=0.18\textwidth]{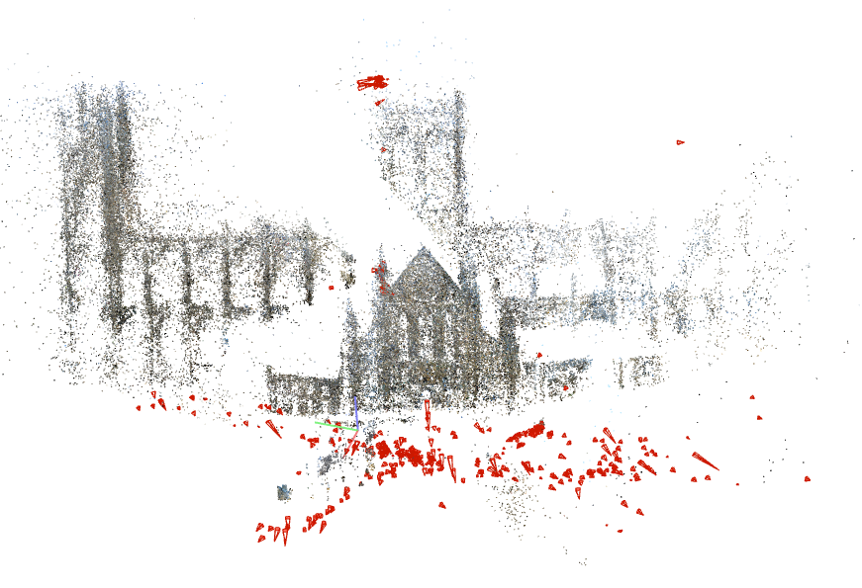}

    \caption{
    Example reconstructions from the proposed \method on various datasets.
    }
    \label{fig:reconstructions_more}
\end{figure}

\noindent\textbf{Qualitative results.} 
Table~\cref{fig:large_scenes_fig} shows 3D reconstructions and camera parameters obtained by \method{} for two scenes with more than 1{,}000 images; in both scenes, we register almost all images. These results demonstrate that our method produces superior reconstructions and effectively handles outliers compared to the baselines. Moreover, \method{} is not limited by the number of images, unlike image-based deep methods such as \vggt{} and \vggsfm{}. Additional qualitative results on diverse scenes are shown in \cref{fig:reconstructions_more}

\noindent\textbf{Runtime.}
Table\ref{tab:runtime} reports runtimes using the identical point tracks produced by our preprocessing. For a fair comparison, the reported runtime for \method{} includes all stages after preprocessing: network inference, test-time fine-tuning, triangulation, and the final robust BA refinement. \method{} is substantially faster than COLMAP, GLOMAP, and Theia, and achieves higher throughput. Importantly, these gains come without sacrificing reconstruction quality: \method{} achieves accuracy and coverage comparable to classical pipelines, demonstrating that learned view-graph pose averaging is efficient at scale.

  \begin{table*}[!h]
      \caption{{\small {\bf Runtime.} Given the same point tracks, we compare the runtime of our proposed method (\method) to \resfm and classical methods,
  including \colmap, \theia, and \glomap. We report $N_r/t$ to normalize runtime with respect to the number of registered images.}}
      \centering
      \label{tab:runtime}
      \rowcolors{3}{rowgray}{white}
      \setlength{\tabcolsep}{3pt}
      \renewcommand{\arraystretch}{1.2}
      \begin{adjustbox}{max width=\textwidth}
      \begin{tabular}{lcc|ccc|ccc|ccc|ccc|ccc}
  \multirow{2}{*}{ Scene } & \multirow{2}{*}{ $N_c$ } & \multirow{2}{*}{ Outliers\% } &
  \multicolumn{3}{c}{\textbf{Ours}}& \multicolumn{3}{c}{\resfm}&
  \multicolumn{3}{c}{\colmap}& \multicolumn{3}{c}{\theia}& \multicolumn{3}{c}{\glomap}\\
  ~ & ~ & ~ &
  Total (Mins) & $N_r$ & $N_r/t\uparrow$
  & Total (Mins) & $N_r$ & $N_r/t\uparrow$
  & Total (Mins) & $N_r$ & $N_r/t\uparrow$
  & Total (Mins) & $N_r$ & $N_r/t\uparrow$
  & Total (Mins) & $N_r$ & $N_r/t\uparrow$ \\
  Alamo                 & 573 & 32.6 & 3.9 & 523 & \textbf{134.1} & 17.2 & 484 & 28.2 & 83.7 & 568 & 6.8  & 13.4 & 553 & 41.4  & 40.0 & 557 & 13.9 \\
  Ellis Island          & 227 & 25.1 & 0.9 & 215 & \textbf{239.3} & 2.8  & 214 & 75.9 & 14.9 & 223 & 15.0 & 1.1  & 213 & 193.6 & 7.7  & 219 & 28.6 \\
  Madrid Metropolis     & 333 & 39.4 & 1.3 & 298 & \textbf{233.4} & 5.8  & 244 & 42.1 & 25.1 & 323 & 12.9 & --   & --  & --    & 7.1  & 320 & 45.2 \\
  Montreal Notre Dame   & 448 & 31.7 & 2.1 & 442 & \textbf{206.4} & 6.1  & 346 & 56.7 & 35.9 & 447 & 12.5 & 3.7  & 422 & 114.6 & 13.5 & 444 & 32.9 \\
  Notre Dame            & 549 & 35.6 & 3.3 & 527 & \textbf{160.2} & 22.2 & 517 & 23.3 & 72.6 & 546 & 7.5  & 11.6 & 534 & 46.0  & 21.1 & 543 & 25.8 \\
  NYC Library           & 330 & 33.6 & 1.9 & 295 & 159.0          & 4.0  & 224 & 55.7 & 26.6 & 330 & 12.4 & 1.5  & 314 & \textbf{204.2} & 7.3  & 323 & 44.5 \\
  Piazza del Popolo     & 336 & 33.1 & 1.0 & 318 & \textbf{304.0} & 2.7  & 249 & 92.6 & 9.6  & 334 & 34.9 & 3.0  & 325 & 108.8 & 5.9  & 331 & 56.0 \\
  Tower of London       & 467 & 27.0 & 3.6 & 457 & 127.9          & 5.9  & 94  & 15.9 & 65.0 & 467 & 7.2  & 3.1  & 448 & \textbf{142.5} & 23.5 & 466 & 19.8 \\
  Vienna Cathedral      & 824 & 31.4 & 7.2 & 763 & \textbf{106.2} & 23.9 & 479 & 20.0 & 98.9 & 824 & 8.3  & 11.2 & 772 & 68.8  & 41.6 & 822 & 19.8 \\
  Yorkminster           & 432 & 29.0 & 2.9 & 402 & \textbf{137.2} & 7.7  & 331 & 42.9 & 31.4 & 419 & 13.3 & 2.9  & 390 & 135.3 & 14.8 & 418 & 28.2 \\
  \midrule
  \textit{Mean}         & --  & --   & 2.8 & 424 & \textbf{180.8} & 9.8  & 318 & 45.3 & 46.4 & 448 & 13.1 & 5.7  & 441 & 117.2 & 18.2 & 444 & 31.5 \\
      \end{tabular}
      \end{adjustbox}
  \end{table*}

\noindent\textbf{Memory scalability.}
Our GNN operates directly on the edge set of the view graph, storing one feature vector per edge rather than materialising a full $N{\times}N$ attention
matrix.
Memory therefore scales as $\mathcal{O}(|\mathcal{E}| \cdot d)$, where $|\mathcal{E}|$ is the number of edges and $d$ is the feature dimension.
For a sparse $k$-nearest-neighbour graph this is $\mathcal{O}(Nk)$, i.e.\ linear in the number of cameras, making the method practical for large scenes.
Even under a complete graph ($|\mathcal{E}|=\mathcal{O}(N^2)$), the architecture remains tractable for the scene sizes encountered in practice, since no dense
pairwise matrix is ever allocated.


\noindent\textbf{Ablations.}
Ablations confirm that each core component of our method is important.
Reducing the number of GNN layers degrades performance, indicating that deeper message passing is beneficial for aggregating global view-graph information.
We also evaluate replacing our message-passing operator with a graph attention mechanism (GATv2), but observe no improvement.
A randomly initialized model performs poorly, while fine-tuning from random initialization improves performance but remains worse than the full method.
This demonstrates the importance of pretraining, which learns representations that generalize across scenes.
Finally, combining pretraining with fine-tuning yields the best results across rotation, translation, and pose AUC.
All results are reported \emph{before} the final BA refinement; see Table~\ref{tab:ablation_rot_trans}.

\begin{table*}[]
  \centering
  \caption{Ablation study reporting rotation, translation, and pose AUC@30 on the validation set, \emph{before} final BA refinement. Higher is better.}
  \label{tab:ablation_rot_trans}
  \footnotesize
  \resizebox{1.0\textwidth}{!}{
  \begin{tabular}{lccc}
    \toprule
     & Rotation AUC@30 (↑) & Translation AUC@30 (↑) & Pose AUC@30 (↑) \\
    \midrule
    2 GNN layers                         & 0.755 & 0.367 & 0.346 \\
    Graph attention (GATv2)              & 0.788 & 0.385 & 0.364 \\
    \midrule
    Pretrained model (without fine-tuning) & 0.789 & 0.375 & 0.358 \\
    Random init                          & 0.316 & 0.024 & 0.012 \\
    Fine-tuning only (random init)        & 0.728 & 0.377 & 0.352 \\
    \midrule
    \textbf{Full model (pretrain + fine-tune)} & \textbf{0.849} & \textbf{0.402} & \textbf{0.389} \\
    \bottomrule
  \end{tabular}
  }
\end{table*}

\clearpage
\section{Conclusion}

We present \method{}, an unsupervised deep \emph{pose-averaging} network for multiview SfM. The design includes a permutation-equivariant pose-averaging module that enforces consistency of pairwise rotations and translation directions through message passing over the view graph. Additional 3D point triangulation and robust BA refinement ensure high accuracy and recover the 3D structure. Across challenging benchmarks (including MegaDepth, 1DSfM), \method{} outperforms deep methods and remains competitive with strong classical pipelines while maintaining high camera coverage. It is also \emph{fast}: on the same point tracks, \method{} is substantially faster than COLMAP and GLOMAP, and modestly faster than Theia, while scaling to large image collections. Finally, a lightweight view re-integration step can optionally recover a portion of the few remaining discarded views with negligible overhead.

\subsubsection{\fontsize{9}{11}\selectfont Acknowledgement.} \fontsize{9}{11}\selectfont Research was partially supported by the MBZUAI-WIS Joint Program for Artificial Intelligence
Research, by Minerva, by the Israeli Council for Higher Education (CHE) via the Weizmann
Data Science Research Center, and by research grants from the Estates of Tully and Michele Plesser and the Anita James
Rosen and Harry Schutzman Foundations.

%
%
\bibliographystyle{splncs04}
\bibliography{main}

\clearpage
\appendix


\section{Implementation details}
\label{sec:App_implementation}
\textbf{Code and data.} Our code and preprocessed data will be made publicly available.

\noindent
\textbf{Framework.} We train and evaluate on NVIDIA A100 GPUs (80\,GB). The implementation uses PyTorch~\cite{paszke2019pytorch} and the Adam optimizer~\cite{kingma2014adam} with gradient normalization.

\noindent
\textbf{Training.} Each epoch iterates over all training scenes. A held-out validation set is used for early stopping. Validation and test evaluations use the complete view graph. Training on MegaDepth takes approximately 7 hours on a single A100. We fix the random seed to 20.

\noindent
\textbf{Architecture details.} The encoder uses 3 edge-conditioned message-passing layers, with both node and edge features represented using 256-dimensional embeddings, and ReLU activations throughout. The camera head $H_{\mathrm{cams}}$ is a 3-layer MLP with hidden dimension 256.

\noindent
\textbf{Hyperparameter search.} We sweep over (1) learning rate $\{10^{-2},\,10^{-3},\,10^{-4}\}$, (2) network width $\{128,\,256,\,512\}$ for the encoder and heads, and (3) number of layers $\{2,\,3,\,4,\,5\}$.

\noindent
\textbf{Bundle adjustment.} We use Ceres Solver~\cite{ceres-solver} with a Huber loss (scale 0.1) for robustness, following \cite{khatib2025resfm}. In each BA round, we cap the number of iterations at 300 or stop earlier on convergence.

\noindent
\textbf{Uncalibrated reconstruction.}
For experiments where camera intrinsics are not provided, we initialize the calibration using a centered principal point and a focal length proportional to the image size, following common SfM practice. These initial intrinsics are used during preprocessing to estimate pairwise relative poses from the matched image pairs. The resulting relative poses are then used as input to \method{}, which predicts the camera extrinsics. In the final reconstruction stage, we jointly refine the camera extrinsics, 3D points, and intrinsic parameters using bundle adjustment. Thus, the approximate calibration is used only to initialize the pipeline, while the final intrinsics are optimized together with the recovered scene geometry. In the released code, we will also provide calibration averaging as an optional preprocessing step, which may improve robustness in uncalibrated settings.

\section{More details on the ablation}

\paragraph{\textbf{Attention-based architecture.}}
For the attention-based ablation, we replace the original message-passing GNN with a GATv2-based \cite{brody2021attentive} network while keeping the overall prediction pipeline unchanged. The input to the network is a graph whose nodes represent cameras and whose pairwise relative-pose measurements are represented by edges. Each relative measurement, consisting of a relative rotation $R_{ij}$ and a normalized translation direction $t_{ij}$, is first mapped to a learned edge embedding using the same encoder as in the main model.

In the variant used in our experiments, we introduce \emph{measurement nodes} in addition to camera nodes. Concretely, for each relative-pose measurement between cameras $i$ and $j$, we create a measurement node initialized from the corresponding edge embedding. The resulting graph is therefore bipartite: each measurement node is connected to its two endpoint camera nodes. The bipartite edge attributes consist of the relative-pose embedding.

Message passing is performed by alternating two attention-based updates within each layer: a camera-to-measurement update and a measurement-to-camera update. Both updates use GATv2 convolutions, allowing the contributions of neighboring nodes to be weighted adaptively through learned attention coefficients conditioned on node and edge features. Each attention block is followed by residual connections, layer normalization, dropout, and a feed-forward MLP, following a pre-normalized transformer-style design. After several such layers, the final camera-node embeddings are passed to the same regression head as in the base model to predict the camera poses.
\section{Constructing point tracks}
\label{sec:App_tracks}
We follow the preprocessing of \cite{khatib2025resfm} to construct point tracks; see their Appendix for full details.

\noindent
\textbf{Runtime protocol for learning-based baselines.}
We report runtimes using a consistent protocol across methods: runtime is measured after the point-track construction stage and therefore does not include the time required to construct point tracks. This is the same convention used for \method{}. For MASt3R, the reported runtime also excludes the pairwise matching stage, which is substantially more expensive than the sparse matching used in our preprocessing. VGGT uses the VGGSfM tracker to construct tracks; in our experiments, this tracking stage has runtime comparable to our track-construction pipeline, with our preprocessing being slightly faster. Therefore, the runtime comparison with VGGT, MASt3R, and VGGSfM follows a fair and consistent protocol, and does not overstate the efficiency of \method{} relative to these learning-based baselines.

\section{Additional results}
\label{app:results}

Here we present the view reintegration results (Table~\ref{tab:post_processing}). 
Tables~\ref{tabapp:megadepth_auc} and~\ref{tabapp:1dsfm_auc} report the AUC (area under the recall curve) scores, computed using the maximum of the relative rotation and translation errors between every image pair, across different thresholds (in degrees), for both the MegaDepth and 1DSfM experiments.

\begin{table}[h]
    \caption{{\small \textbf{MegaDepth: effect of view reintegration.} 
    We report the number of input images ($N_c$), outlier fraction, registered images ($N_r$), and mean rotation and translation errors for our method (\emph{Ours}) and with the add-back postprocessing step (\emph{Ours + Post-Processing}).}}

    \label{tab:post_processing}
    \centering
    \scriptsize 
    \rowcolors{3}{rowgray}{white} 
    \renewcommand{\arraystretch}{0.7}
    \begin{adjustbox}{max width=\textwidth}
    \begin{tabular}{ccc|ccc|ccc}
\multirow{2}{*}{ Scene } &
\multirow{2}{*}{ $N_c$ } &
\multirow{2}{*}{ Outliers\% } &
\multicolumn{3}{c}{\textbf{Ours}}&
\multicolumn{3}{c}{\textbf{Ours + postprocessing}}\\
 &  &  & $N_r$ & Rot  & Trans   & $N_r$ & Rot  & Trans \\
\midrule

0238 & 522 & $ 44.6\% $  & 486  & $\mathbf{\underline{1.06}}$  & $\mathbf{\underline{0.285}}$  & $\mathbf{\underline{515}}$  & 1.08  & 0.288 \\
0060 & 528 & $ 41.6\% $  & 518  & $\mathbf{\underline{0.10}}$  & $\mathbf{\underline{0.026}}$  & $\mathbf{\underline{526}}$  & $\mathbf{\underline{0.10}}$  & 0.027 \\
0197 & 870 & $ 40.7\% $  & 718  & $\mathbf{\underline{0.58}}$  & $\mathbf{\underline{0.108}}$  & $\mathbf{\underline{844}}$  & 0.89  & 0.151 \\
0094 & 763 & $ 40.1\% $  & 659  & $\mathbf{\underline{0.81}}$  & $\mathbf{\underline{0.116}}$  & $\mathbf{\underline{717}}$  & 1.82  & 0.200 \\
0265 & 571 & $ 38.8\% $  & 372  & $\mathbf{\underline{5.30}}$  & $\mathbf{\underline{1.433}}$  & $\mathbf{\underline{531}}$  & 6.42  & 1.900 \\
0083 & 635 & $ 31.3\% $  & 622  & $\mathbf{\underline{0.32}}$  & $\mathbf{\underline{0.062}}$  & $\mathbf{\underline{632}}$  & $\mathbf{\underline{0.32}}$  & $\mathbf{\underline{0.062}}$ \\
0076 & 558 & $ 30.5\% $  & 547  & $\mathbf{\underline{0.09}}$  & 0.037  & $\mathbf{\underline{552}}$  & $\mathbf{\underline{0.09}}$  & $\mathbf{\underline{0.036}}$ \\
0185 & 368 & $ 30.0\% $  & 364  & $\mathbf{\underline{0.12}}$  & $\mathbf{\underline{0.021}}$  & $\mathbf{\underline{367}}$  & 0.12  & $\mathbf{\underline{0.021}}$ \\
0048 & 512 & $ 24.2\% $  & 501  & $\mathbf{\underline{0.10}}$  & $\mathbf{\underline{0.011}}$  & $\mathbf{\underline{507}}$  & 0.10  & $\mathbf{\underline{0.011}}$ \\
0024 & 356 & $ 23.0\% $  & 328  & $\mathbf{\underline{3.59}}$  & $\mathbf{\underline{1.122}}$  & $\mathbf{\underline{349}}$  & 3.81  & 1.142 \\
0223 & 214 & $ 17.0\% $  & 211  & $\mathbf{\underline{3.45}}$  & $\mathbf{\underline{0.285}}$  & $\mathbf{\underline{214}}$  & 3.55  & 0.291 \\
5016 & 28 & $ 16.9\% $  & $\mathbf{\underline{28}}$  & $\mathbf{\underline{0.08}}$  & $\mathbf{\underline{0.015}}$  & $\mathbf{\underline{28}}$  & $\mathbf{\underline{0.08}}$  & $\mathbf{\underline{0.015}}$ \\
0046 & 440 & $ 14.6\% $  & 438  & 0.47  & $\mathbf{\underline{0.073}}$  & $\mathbf{\underline{440}}$  & $\mathbf{\underline{0.46}}$  & $\mathbf{\underline{0.073}}$ \\
\midrule \midrule
0099 & 299 & $ 47.4\% $  & 157  & $\mathbf{\underline{0.56}}$  & $\mathbf{\underline{0.229}}$  & $\mathbf{\underline{215}}$  & 0.93  & 0.291 \\
1001 & 285 & $ 43.9\% $  & 268  & 3.87  & $\mathbf{\underline{2.585}}$  & $\mathbf{\underline{277}}$  & $\mathbf{\underline{3.78}}$  & 2.595 \\
0231 & 296 & $ 42.2\% $  & 260  & $\mathbf{\underline{0.31}}$  & $\mathbf{\underline{0.029}}$  & $\mathbf{\underline{279}}$  & 0.52  & 0.063 \\
0411 & 299 & $ 29.9\% $  & 268  & $\mathbf{\underline{0.23}}$  & $\mathbf{\underline{0.036}}$  & $\mathbf{\underline{292}}$  & 0.26  & 0.044 \\
0377 & 295 & $ 27.5\% $  & 232  & $\mathbf{\underline{0.12}}$  & $\mathbf{\underline{0.016}}$  & $\mathbf{\underline{263}}$  & 0.18  & 0.021 \\
0102 & 299 & $ 25.8\% $  & 296  & $\mathbf{\underline{0.18}}$  & $\mathbf{\underline{0.023}}$  & $\mathbf{\underline{298}}$  & 0.18  & $\mathbf{\underline{0.023}}$ \\
0147 & 298 & $ 24.6\% $  & 289  & $\mathbf{\underline{1.36}}$  & $\mathbf{\underline{0.118}}$  & $\mathbf{\underline{295}}$  & 1.84  & 0.140 \\
0148 & 287 & $ 24.6\% $  & 244  & $\mathbf{\underline{1.15}}$  & $\mathbf{\underline{0.135}}$  & $\mathbf{\underline{262}}$  & 2.58  & 0.270 \\
0446 & 298 & $ 22.1\% $  & 296  & $\mathbf{\underline{0.34}}$  & $\mathbf{\underline{0.023}}$  & $\mathbf{\underline{298}}$  & 0.35  & 0.024 \\
0022 & 297 & $ 21.2\% $  & 280  & $\mathbf{\underline{0.11}}$  & $\mathbf{\underline{0.015}}$  & $\mathbf{\underline{288}}$  & 0.11  & 0.016 \\
0327 & 298 & $ 21.0\% $  & 280  & $\mathbf{\underline{0.48}}$  & $\mathbf{\underline{0.032}}$  & $\mathbf{\underline{293}}$  & 1.45  & 0.085 \\
0015 & 284 & $ 20.6\% $  & 241  & $\mathbf{\underline{0.61}}$  & $\mathbf{\underline{0.083}}$  & $\mathbf{\underline{260}}$  & 0.89  & 0.153 \\
0455 & 298 & $ 19.8\% $  & 292  & $\mathbf{\underline{0.40}}$  & $\mathbf{\underline{0.071}}$  & $\mathbf{\underline{298}}$  & 0.41  & $\mathbf{\underline{0.071}}$ \\
0496 & 297 & $ 19.2\% $  & 280  & $\mathbf{\underline{0.69}}$  & $\mathbf{\underline{0.041}}$  & $\mathbf{\underline{292}}$  & 0.70  & 0.047 \\
1589 & 299 & $ 17.4\% $  & 294  & $\mathbf{\underline{0.13}}$  & $\mathbf{\underline{0.074}}$  & $\mathbf{\underline{298}}$  & 0.14  & 0.075 \\
0012 & 299 & $ 16.3\% $  & 295  & $\mathbf{\underline{0.77}}$  & $\mathbf{\underline{0.086}}$  & $\mathbf{\underline{298}}$  & 1.18  & 0.137 \\
0104 & 284 & $ 16.2\% $  & 237  & $\mathbf{\underline{4.04}}$  & $\mathbf{\underline{0.330}}$  & $\mathbf{\underline{256}}$  & 9.01  & 0.707 \\
0019 & 299 & $ 15.4\% $  & 288  & $\mathbf{\underline{0.23}}$  & $\mathbf{\underline{0.008}}$  & $\mathbf{\underline{296}}$  & 0.32  & 0.012 \\
0063 & 293 & $ 14.5\% $  & 266  & $\mathbf{\underline{0.14}}$  & $\mathbf{\underline{0.024}}$  & $\mathbf{\underline{278}}$  & 0.32  & 0.035 \\
0130 & 285 & $ 14.4\% $  & 202  & $\mathbf{\underline{0.21}}$  & $\mathbf{\underline{0.019}}$  & $\mathbf{\underline{212}}$  & 0.87  & 0.099 \\
0080 & 284 & $ 12.9\% $  & 141  & $\mathbf{\underline{0.10}}$  & $\mathbf{\underline{0.017}}$  & $\mathbf{\underline{159}}$  & 1.93  & 0.231 \\
0240 & 298 & $ 11.9\% $  & 288  & 0.64  & 0.088  & $\mathbf{\underline{296}}$  & $\mathbf{\underline{0.62}}$  & $\mathbf{\underline{0.087}}$ \\
0007 & 290 & $ 11.7\% $  & 284  & 1.51  & 0.079  & $\mathbf{\underline{289}}$  & $\mathbf{\underline{1.49}}$  & $\mathbf{\underline{0.077}}$ \\
Mean & 379 & $ 26.1\% $  & 327  & $\mathbf{\underline{0.95}}$  & $\mathbf{\underline{0.215}}$  & $\mathbf{\underline{348}}$  & 1.36  & 0.264 \\

    \end{tabular}
    \end{adjustbox}
\end{table}

In Tables ~\ref{tabapp:megadepth_ra_auc} and~\ref{tabapp:1dsfm_ra_auc} we additionally report a registration-aware AUC (RA-AUC), which accounts for both pose accuracy and camera registration coverage. 
Unlike the standard AUC computed only over registered cameras, RA-AUC is computed over all $\binom{N_c}{2}$ image pairs. 
For any pair containing at least one unregistered image, we assign an error of $180^\circ$, so that the pair contributes zero recall at all evaluated thresholds. 
Equivalently, if the original AUC is computed only over registered image pairs, the registration-aware score is
\[
\mathrm{RA\text{-}AUC}
=
\mathrm{AUC}
\cdot
\frac{\binom{N_r}{2}}{\binom{N_c}{2}} .
\]
This metric penalizes incomplete reconstructions and therefore better reflects the trade-off between accuracy and registration completeness.

\noindent
\textbf{Note on \resfm AUC results.} To compute the AUC and registration-aware AUC metrics, we re-ran \resfm using the same evaluation protocol as the other methods. Since the final bundle-adjustment stage is not fully deterministic, the resulting \resfm numbers may differ slightly from the original numbers reported in the \resfm paper and used in our main paper tables. These differences are minor and do not affect the conclusions.

\clearpage

  \begin{table*}[t]
  \caption{{\small {\bf MegaDepth experiment (AUC).}
  For each scene, we list the number of input images ($N_c$) and the fraction of outliers.
  For each model, we report the number of registered cameras ($N_r$) and the AUC values at different error thresholds (in degrees). Winning results are marked in
  \textbf{\underline{bold and underlined}}.}}
      \label{tabapp:megadepth_auc}
      \centering
      \tiny
      \rowcolors{3}{rowgray}{white}
      \setlength{\tabcolsep}{3pt}
      \renewcommand{\arraystretch}{1.2}
      \begin{adjustbox}{max width=\textwidth}
      \begin{tabular}{lcc|cccc|cccc|cccc|cccc}

  \multirow{2}{*}{ Scene } &
  \multirow{2}{*}{ $N_c$ } &
  \multirow{2}{*}{ Outliers\% } &
  \multicolumn{4}{c|}{\textbf{Ours}}&
  \multicolumn{4}{c|}{RESFM} &
  \multicolumn{4}{c|}{\theia}&
  \multicolumn{4}{c}{\glomap}\\

   &  &   & $N_r$ & @1 & @5 & @30 & $N_r$ & @1 & @5 & @30 & $N_r$ & @1 & @5 & @30 & $N_r$ & @1 & @5 & @30 \\

0238 & 522 & $ 44.6\% $  & 486  & 0.427  & 0.648  & 0.887  & 354  & $\mathbf{\underline{0.547}}$  & $\mathbf{\underline{0.811}}$  & $\mathbf{\underline{0.916}}$  & $\mathbf{\underline{506}}$  & 0.063  & 0.495  & 0.863  & 497  & 0.298  & 0.653  & 0.884 \\
0060 & 528 & $ 41.6\% $  & 518  & $\mathbf{\underline{0.811}}$  & $\mathbf{\underline{0.935}}$  & $\mathbf{\underline{0.982}}$  & 491  & 0.762  & 0.906  & 0.969  & $\mathbf{\underline{525}}$  & 0.300  & 0.702  & 0.912  & 520  & 0.676  & 0.893  & 0.969 \\
0197 & 870 & $ 40.7\% $  & 718  & 0.394  & 0.771  & 0.944  & 706  & $\mathbf{\underline{0.643}}$  & $\mathbf{\underline{0.865}}$  & $\mathbf{\underline{0.954}}$  & $\mathbf{\underline{855}}$  & 0.086  & 0.505  & 0.872  & 813  & 0.555  & 0.817  & 0.940 \\
0094 & 763 & $ 40.1\% $  & 659  & $\mathbf{\underline{0.704}}$  & $\mathbf{\underline{0.879}}$  & $\mathbf{\underline{0.949}}$  & 539  & 0.365  & 0.533  & 0.749  & $\mathbf{\underline{742}}$  & 0.325  & 0.708  & 0.904  & 711  & 0.468  & 0.772  & 0.921 \\
0265 & 571 & $ 38.8\% $  & 372  & 0.000  & 0.004  & 0.181  & 457  & $\mathbf{\underline{0.129}}$  & $\mathbf{\underline{0.641}}$  & $\mathbf{\underline{0.911}}$  & $\mathbf{\underline{554}}$  & 0.000  & 0.005  & 0.267  & $\mathbf{\underline{554}}$  & 0.000  & 0.001  & 0.165 \\
0083 & 635 & $ 31.3\% $  & 622  & 0.852  & 0.945  & 0.981  & 572  & $\mathbf{\underline{0.855}}$  & $\mathbf{\underline{0.959}}$  & $\mathbf{\underline{0.989}}$  & $\mathbf{\underline{632}}$  & 0.504  & 0.817  & 0.948  & 614  & 0.765  & 0.935  & 0.987 \\
0076 & 558 & $ 30.5\% $  & 547  & $\mathbf{\underline{0.765}}$  & 0.925  & 0.980  & 512  & 0.732  & $\mathbf{\underline{0.931}}$  & $\mathbf{\underline{0.986}}$  & $\mathbf{\underline{549}}$  & 0.133  & 0.599  & 0.897  & 540  & 0.510  & 0.830  & 0.963 \\
0185 & 368 & $ 30.0\% $  & 364  & $\mathbf{\underline{0.847}}$  & 0.954  & 0.985  & 345  & 0.812  & $\mathbf{\underline{0.956}}$  & $\mathbf{\underline{0.992}}$  & $\mathbf{\underline{365}}$  & 0.285  & 0.742  & 0.937  & $\mathbf{\underline{365}}$  & 0.641  & 0.889  & 0.970 \\
0048 & 512 & $ 24.2\% $  & 501  & $\mathbf{\underline{0.838}}$  & $\mathbf{\underline{0.955}}$  & $\mathbf{\underline{0.991}}$  & 485  & 0.800  & 0.941  & 0.976  & $\mathbf{\underline{507}}$  & 0.397  & 0.784  & 0.949  & 505  & 0.698  & 0.907  & 0.975 \\
0024 & 356 & $ 23.0\% $  & 328  & 0.513  & 0.749  & 0.827  & 308  & $\mathbf{\underline{0.522}}$  & $\mathbf{\underline{0.829}}$  & $\mathbf{\underline{0.948}}$  & $\mathbf{\underline{355}}$  & 0.153  & 0.555  & 0.873  & 338  & 0.362  & 0.719  & 0.931 \\
0223 & 214 & $ 17.0\% $  & 211  & $\mathbf{\underline{0.599}}$  & $\mathbf{\underline{0.827}}$  & $\mathbf{\underline{0.916}}$  & 202  & 0.483  & 0.756  & 0.864  & 212  & 0.014  & 0.342  & 0.783  & $\mathbf{\underline{213}}$  & 0.330  & 0.703  & 0.908 \\
5016 & 28 & $ 16.9\% $  & $\mathbf{\underline{28}}$  & $\mathbf{\underline{0.790}}$  & $\mathbf{\underline{0.928}}$  & $\mathbf{\underline{0.984}}$  & $\mathbf{\underline{28}}$  & 0.355  & 0.499  & 0.741  & $\mathbf{\underline{28}}$  & 0.413  & 0.793  & 0.952  & $\mathbf{\underline{28}}$  & 0.508  & 0.835  & 0.960 \\
0046 & 440 & $ 14.6\% $  & 438  & $\mathbf{\underline{0.928}}$  & 0.971  & 0.985  & 388  & 0.607  & 0.687  & 0.737  & 434  & 0.530  & 0.861  & 0.965  & $\mathbf{\underline{440}}$  & 0.895  & $\mathbf{\underline{0.977}}$  & $\mathbf{\underline{0.996}}$ \\
\midrule \midrule
0099 & 299 & $ 47.4\% $  & 157  & 0.239  & 0.637  & 0.869  & 222  & 0.048  & 0.319  & 0.636  & $\mathbf{\underline{297}}$  & 0.011  & 0.172  & 0.619  & 255  & $\mathbf{\underline{0.526}}$  & $\mathbf{\underline{0.769}}$  & $\mathbf{\underline{0.909}}$ \\
1001 & 285 & $ 43.9\% $  & 268  & $\mathbf{\underline{0.018}}$  & $\mathbf{\underline{0.314}}$  & 0.697  & 256  & 0.005  & 0.241  & $\mathbf{\underline{0.779}}$  & $\mathbf{\underline{275}}$  & 0.000  & 0.001  & 0.051  & 270  & 0.000  & 0.003  & 0.133 \\
0231 & 296 & $ 42.2\% $  & 260  & $\mathbf{\underline{0.599}}$  & $\mathbf{\underline{0.867}}$  & $\mathbf{\underline{0.968}}$  & 247  & 0.542  & 0.855  & 0.954  & $\mathbf{\underline{286}}$  & 0.063  & 0.467  & 0.842  & 278  & 0.417  & 0.762  & 0.921 \\
0411 & 299 & $ 29.9\% $  & 268  & 0.692  & 0.910  & 0.977  & 273  & $\mathbf{\underline{0.693}}$  & $\mathbf{\underline{0.917}}$  & $\mathbf{\underline{0.984}}$  & $\mathbf{\underline{293}}$  & 0.188  & 0.600  & 0.902  & 268  & 0.380  & 0.727  & 0.928 \\
0377 & 295 & $ 27.5\% $  & 232  & $\mathbf{\underline{0.793}}$  & $\mathbf{\underline{0.943}}$  & 0.986  & 197  & 0.769  & 0.938  & $\mathbf{\underline{0.989}}$  & $\mathbf{\underline{269}}$  & 0.198  & 0.596  & 0.882  & 266  & 0.567  & 0.824  & 0.941 \\
0102 & 299 & $ 25.8\% $  & $\mathbf{\underline{296}}$  & $\mathbf{\underline{0.763}}$  & $\mathbf{\underline{0.922}}$  & $\mathbf{\underline{0.979}}$  & 280  & 0.679  & 0.906  & 0.977  & 294  & 0.169  & 0.474  & 0.785  & 293  & 0.547  & 0.805  & 0.946 \\
0147 & 298 & $ 24.6\% $  & $\mathbf{\underline{289}}$  & $\mathbf{\underline{0.635}}$  & $\mathbf{\underline{0.778}}$  & $\mathbf{\underline{0.924}}$  & 241  & 0.416  & 0.552  & 0.694  & 284  & 0.055  & 0.468  & 0.771  & $\mathbf{\underline{289}}$  & 0.000  & 0.000  & 0.000 \\
0148 & 287 & $ 24.6\% $  & 244  & $\mathbf{\underline{0.504}}$  & $\mathbf{\underline{0.698}}$  & $\mathbf{\underline{0.926}}$  & 256  & 0.416  & 0.546  & 0.672  & 275  & 0.050  & 0.280  & 0.499  & $\mathbf{\underline{282}}$  & 0.296  & 0.472  & 0.591 \\
0446 & 298 & $ 22.1\% $  & $\mathbf{\underline{296}}$  & $\mathbf{\underline{0.677}}$  & $\mathbf{\underline{0.896}}$  & 0.972  & 288  & 0.627  & 0.891  & $\mathbf{\underline{0.973}}$  & 289  & 0.053  & 0.460  & 0.840  & 295  & 0.465  & 0.799  & 0.947 \\
0022 & 297 & $ 21.2\% $  & 280  & $\mathbf{\underline{0.706}}$  & $\mathbf{\underline{0.925}}$  & $\mathbf{\underline{0.987}}$  & 273  & 0.676  & 0.916  & 0.985  & $\mathbf{\underline{296}}$  & 0.194  & 0.631  & 0.900  & 280  & 0.473  & 0.797  & 0.949 \\
0327 & 298 & $ 21.0\% $  & 280  & $\mathbf{\underline{0.767}}$  & 0.922  & 0.974  & 270  & 0.751  & $\mathbf{\underline{0.931}}$  & $\mathbf{\underline{0.983}}$  & $\mathbf{\underline{288}}$  & 0.040  & 0.538  & 0.884  & $\mathbf{\underline{288}}$  & 0.474  & 0.683  & 0.776 \\
0015 & 284 & $ 20.6\% $  & 241  & $\mathbf{\underline{0.767}}$  & $\mathbf{\underline{0.896}}$  & $\mathbf{\underline{0.951}}$  & 223  & 0.547  & 0.718  & 0.808  & 244  & 0.134  & 0.478  & 0.784  & $\mathbf{\underline{272}}$  & 0.572  & 0.839  & 0.950 \\
0455 & 298 & $ 19.8\% $  & 292  & $\mathbf{\underline{0.735}}$  & 0.892  & 0.961  & 293  & 0.712  & $\mathbf{\underline{0.916}}$  & $\mathbf{\underline{0.979}}$  & 294  & 0.209  & 0.644  & 0.905  & $\mathbf{\underline{297}}$  & 0.585  & 0.862  & 0.966 \\
0496 & 297 & $ 19.2\% $  & 280  & $\mathbf{\underline{0.721}}$  & $\mathbf{\underline{0.909}}$  & $\mathbf{\underline{0.966}}$  & 281  & 0.615  & 0.876  & 0.963  & 285  & 0.080  & 0.498  & 0.849  & $\mathbf{\underline{291}}$  & 0.441  & 0.786  & 0.942 \\
1589 & 299 & $ 17.4\% $  & 294  & 0.654  & 0.898  & 0.971  & 290  & $\mathbf{\underline{0.744}}$  & $\mathbf{\underline{0.930}}$  & $\mathbf{\underline{0.987}}$  & 288  & 0.157  & 0.489  & 0.797  & $\mathbf{\underline{298}}$  & 0.583  & 0.812  & 0.951 \\
0012 & 299 & $ 16.3\% $  & $\mathbf{\underline{295}}$  & $\mathbf{\underline{0.786}}$  & $\mathbf{\underline{0.907}}$  & 0.957  & 291  & 0.682  & 0.825  & 0.862  & 129  & 0.087  & 0.499  & 0.840  & $\mathbf{\underline{295}}$  & 0.645  & 0.887  & $\mathbf{\underline{0.971}}$ \\
0104 & 284 & $ 16.2\% $  & 237  & 0.590  & 0.698  & 0.846  & 195  & $\mathbf{\underline{0.727}}$  & $\mathbf{\underline{0.932}}$  & $\mathbf{\underline{0.988}}$  & 265  & 0.127  & 0.430  & 0.642  & $\mathbf{\underline{280}}$  & 0.445  & 0.634  & 0.717 \\
0019 & 299 & $ 15.4\% $  & 288  & $\mathbf{\underline{0.855}}$  & 0.950  & 0.987  & 247  & 0.832  & $\mathbf{\underline{0.962}}$  & $\mathbf{\underline{0.993}}$  & 271  & 0.243  & 0.649  & 0.888  & $\mathbf{\underline{296}}$  & 0.740  & 0.924  & 0.986 \\
0063 & 293 & $ 14.5\% $  & 266  & $\mathbf{\underline{0.752}}$  & $\mathbf{\underline{0.936}}$  & $\mathbf{\underline{0.988}}$  & 263  & 0.704  & 0.922  & 0.986  & 268  & 0.106  & 0.526  & 0.879  & $\mathbf{\underline{287}}$  & 0.460  & 0.805  & 0.956 \\
0130 & 285 & $ 14.4\% $  & 202  & $\mathbf{\underline{0.770}}$  & 0.910  & $\mathbf{\underline{0.983}}$  & 193  & 0.714  & $\mathbf{\underline{0.911}}$  & 0.969  & 187  & 0.057  & 0.424  & 0.828  & $\mathbf{\underline{279}}$  & 0.347  & 0.549  & 0.835 \\
0080 & 284 & $ 12.9\% $  & 141  & $\mathbf{\underline{0.757}}$  & $\mathbf{\underline{0.935}}$  & $\mathbf{\underline{0.987}}$  & 140  & 0.709  & 0.909  & 0.972  & 278  & 0.035  & 0.278  & 0.720  & $\mathbf{\underline{282}}$  & 0.259  & 0.512  & 0.905 \\
0240 & 298 & $ 11.9\% $  & 288  & $\mathbf{\underline{0.713}}$  & $\mathbf{\underline{0.917}}$  & $\mathbf{\underline{0.974}}$  & 276  & 0.599  & 0.837  & 0.922  & 285  & 0.141  & 0.536  & 0.867  & $\mathbf{\underline{290}}$  & 0.367  & 0.709  & 0.923 \\
0007 & 290 & $ 11.7\% $  & 284  & $\mathbf{\underline{0.791}}$  & 0.897  & 0.948  & 172  & 0.716  & $\mathbf{\underline{0.906}}$  & 0.973  & 277  & 0.069  & 0.591  & 0.902  & $\mathbf{\underline{289}}$  & 0.659  & 0.881  & $\mathbf{\underline{0.974}}$ \\
Mean & 379 & $ 26.1\% $  & 327  & $\mathbf{\underline{0.660}}$  & $\mathbf{\underline{0.835}}$  & $\mathbf{\underline{0.927}}$  & 307  & 0.598  & 0.805  & 0.910  & 347  & 0.157  & 0.518  & 0.805  & $\mathbf{\underline{352}}$  & 0.471  & 0.716  & 0.852 \\

    \end{tabular}
    \end{adjustbox}

\end{table*}

\vspace{-2mm}

  \begin{table*}[]
  \caption{{\small {\bf 1DSfM experiment (AUC).}
  For each scene, we list the number of input images ($N_c$) and the fraction of outliers.
  For each model, we report the number of registered cameras ($N_r$) and the AUC values at different error thresholds (in degrees). Winning results are marked in
  \textbf{\underline{bold and underlined}}.}}

      \label{tabapp:1dsfm_auc}
      \centering
      \tiny
      \rowcolors{3}{rowgray}{white}
      \setlength{\tabcolsep}{3pt}
      \renewcommand{\arraystretch}{1.2}
      \begin{adjustbox}{max width=\textwidth}
      \begin{tabular}{lcc|cccc|cccc|cccc|cccc}

  \multirow{2}{*}{ Scene } &
  \multirow{2}{*}{ $N_c$ } &
  \multirow{2}{*}{ Outliers\% } &
  \multicolumn{4}{c|}{\textbf{Ours}}&
  \multicolumn{4}{c|}{RESFM} &
  \multicolumn{4}{c|}{\theia}&
  \multicolumn{4}{c}{\glomap}\\

   &  &   & $N_r$ & @1 & @5 & @30 & $N_r$ & @1 & @5 & @30 & $N_r$ & @1 & @5 & @30 & $N_r$ & @1 & @5 & @30 \\

Alamo & 573 & $ 32.6\% $  & 523  & $\mathbf{\underline{0.430}}$  & $\mathbf{\underline{0.671}}$  & $\mathbf{\underline{0.841}}$  & 496  & 0.409  & 0.668  & 0.784  & 549  & 0.003  & 0.093  & 0.499  & $\mathbf{\underline{557}}$  & 0.092  & 0.482  & 0.833 \\
Ellis Island & 227 & $ 25.1\% $  & 215  & $\mathbf{\underline{0.421}}$  & $\mathbf{\underline{0.841}}$  & $\mathbf{\underline{0.964}}$  & 211  & 0.285  & 0.766  & 0.927  & 213  & 0.000  & 0.023  & 0.441  & $\mathbf{\underline{219}}$  & 0.071  & 0.539  & 0.901 \\
Madrid Metropolis & 333 & $ 39.4\% $  & 298  & $\mathbf{\underline{0.590}}$  & $\mathbf{\underline{0.819}}$  & $\mathbf{\underline{0.944}}$  & 247  & 0.342  & 0.560  & 0.701  & 319  & 0.015  & 0.241  & 0.650  & $\mathbf{\underline{320}}$  & 0.163  & 0.626  & 0.876 \\
Montreal Notre Dame & 448 & $ 31.7\% $  & 442  & 0.503  & 0.787  & 0.894  & 333  & $\mathbf{\underline{0.528}}$  & $\mathbf{\underline{0.790}}$  & $\mathbf{\underline{0.933}}$  & 420  & 0.001  & 0.093  & 0.553  & $\mathbf{\underline{444}}$  & 0.090  & 0.549  & 0.890 \\
NYC Library & 330 & $ 33.6\% $  & 295  & $\mathbf{\underline{0.580}}$  & $\mathbf{\underline{0.852}}$  & $\mathbf{\underline{0.959}}$  & 281  & 0.286  & 0.521  & 0.620  & 313  & 0.006  & 0.161  & 0.566  & $\mathbf{\underline{323}}$  & 0.141  & 0.634  & 0.910 \\
Notre Dame & 549 & $ 35.6\% $  & 527  & $\mathbf{\underline{0.426}}$  & $\mathbf{\underline{0.775}}$  & $\mathbf{\underline{0.951}}$  & 510  & 0.378  & 0.729  & 0.918  & 531  & 0.015  & 0.293  & 0.726  & $\mathbf{\underline{543}}$  & 0.101  & 0.566  & 0.864 \\
Piazza del Popolo & 336 & $ 33.1\% $  & 318  & 0.435  & 0.602  & 0.757  & 248  & $\mathbf{\underline{0.593}}$  & $\mathbf{\underline{0.802}}$  & 0.893  & 324  & 0.025  & 0.226  & 0.608  & $\mathbf{\underline{331}}$  & 0.203  & 0.648  & $\mathbf{\underline{0.899}}$ \\
Tower of London & 467 & $ 27.0\% $  & 457  & $\mathbf{\underline{0.487}}$  & $\mathbf{\underline{0.820}}$  & $\mathbf{\underline{0.945}}$  & 116  & 0.237  & 0.568  & 0.713  & 448  & 0.002  & 0.093  & 0.474  & $\mathbf{\underline{466}}$  & 0.114  & 0.600  & 0.897 \\
Vienna Cathedral & 824 & $ 31.4\% $  & 763  & $\mathbf{\underline{0.371}}$  & $\mathbf{\underline{0.586}}$  & 0.709  & 578  & 0.340  & 0.584  & 0.706  & 767  & 0.000  & 0.008  & 0.270  & $\mathbf{\underline{822}}$  & 0.053  & 0.499  & $\mathbf{\underline{0.846}}$ \\
Yorkminster & 432 & $ 29.0\% $  & 402  & $\mathbf{\underline{0.512}}$  & $\mathbf{\underline{0.849}}$  & $\mathbf{\underline{0.962}}$  & 304  & 0.080  & 0.404  & 0.628  & 387  & 0.000  & 0.038  & 0.357  & $\mathbf{\underline{418}}$  & 0.103  & 0.600  & 0.905 \\
Mean & 451 & $ 31.9\% $  & 424  & $\mathbf{\underline{0.476}}$  & $\mathbf{\underline{0.760}}$  & $\mathbf{\underline{0.893}}$  & 332  & 0.348  & 0.639  & 0.782  & 427  & 0.007  & 0.127  & 0.514  & $\mathbf{\underline{444}}$  & 0.113  & 0.574  & 0.882 \\

    \end{tabular}
    \end{adjustbox}

\end{table*}

\clearpage

  \begin{table*}[t]
\caption{{\small {\bf MegaDepth experiment (registration-aware AUC).}
For each scene, we list the number of input images ($N_c$) and the fraction of outliers.
For each model, we report the number of registered cameras ($N_r$) and the registration-aware AUC (RA-AUC) values at different error thresholds (in degrees).
RA-AUC is computed over all image pairs; pairs involving at least one unregistered camera are assigned an error of $180^\circ$ and therefore contribute zero recall at all reported thresholds.
Winning results are marked in \textbf{\underline{bold and underlined}}.}}
      \label{tabapp:megadepth_ra_auc}
      \centering
      \tiny
      \rowcolors{3}{rowgray}{white}
      \setlength{\tabcolsep}{3pt}
      \renewcommand{\arraystretch}{1.2}
      \begin{adjustbox}{max width=\textwidth}
      \begin{tabular}{lcc|cccc|cccc|cccc|cccc}

  \multirow{2}{*}{ Scene } &
  \multirow{2}{*}{ $N_c$ } &
  \multirow{2}{*}{ Outliers\% } &
  \multicolumn{4}{c|}{\textbf{Ours}}&
  \multicolumn{4}{c|}{RESFM} &
  \multicolumn{4}{c|}{\theia}&
  \multicolumn{4}{c}{\glomap}\\

   &  &   & $N_r$ & @1 & @5 & @30 & $N_r$ & @1 & @5 & @30 & $N_r$ & @1 & @5 & @30 & $N_r$ & @1 & @5 & @30 \\

0238 & 522 & $ 44.6\% $  & 486  & $\mathbf{\underline{0.370}}$  & 0.562  & 0.769  & 354  & 0.251  & 0.373  & 0.421  & $\mathbf{\underline{506}}$  & 0.059  & 0.466  & $\mathbf{\underline{0.810}}$  & 497  & 0.270  & $\mathbf{\underline{0.592}}$  & 0.801 \\
0060 & 528 & $ 41.6\% $  & 518  & $\mathbf{\underline{0.780}}$  & $\mathbf{\underline{0.900}}$  & $\mathbf{\underline{0.945}}$  & 491  & 0.659  & 0.783  & 0.838  & $\mathbf{\underline{525}}$  & 0.296  & 0.694  & 0.901  & 520  & 0.656  & 0.866  & 0.940 \\
0197 & 870 & $ 40.7\% $  & 718  & 0.268  & 0.525  & 0.643  & 706  & 0.423  & 0.570  & 0.628  & $\mathbf{\underline{855}}$  & 0.083  & 0.487  & $\mathbf{\underline{0.842}}$  & 813  & $\mathbf{\underline{0.484}}$  & $\mathbf{\underline{0.714}}$  & 0.821 \\
0094 & 763 & $ 40.1\% $  & 659  & $\mathbf{\underline{0.525}}$  & 0.656  & 0.708  & 539  & 0.182  & 0.266  & 0.374  & $\mathbf{\underline{742}}$  & 0.307  & 0.669  & $\mathbf{\underline{0.855}}$  & 711  & 0.406  & $\mathbf{\underline{0.670}}$  & 0.800 \\
0265 & 571 & $ 38.8\% $  & 372  & 0.000  & 0.002  & 0.077  & 457  & $\mathbf{\underline{0.083}}$  & $\mathbf{\underline{0.410}}$  & $\mathbf{\underline{0.583}}$  & $\mathbf{\underline{554}}$  & 0.000  & 0.005  & 0.251  & $\mathbf{\underline{554}}$  & 0.000  & 0.001  & 0.155 \\
0083 & 635 & $ 31.3\% $  & 622  & $\mathbf{\underline{0.817}}$  & $\mathbf{\underline{0.907}}$  & $\mathbf{\underline{0.941}}$  & 572  & 0.694  & 0.778  & 0.802  & $\mathbf{\underline{632}}$  & 0.499  & 0.810  & 0.939  & 614  & 0.715  & 0.874  & 0.923 \\
0076 & 558 & $ 30.5\% $  & 547  & $\mathbf{\underline{0.735}}$  & $\mathbf{\underline{0.889}}$  & $\mathbf{\underline{0.942}}$  & 512  & 0.616  & 0.784  & 0.830  & $\mathbf{\underline{549}}$  & 0.129  & 0.580  & 0.869  & 540  & 0.478  & 0.777  & 0.901 \\
0185 & 368 & $ 30.0\% $  & 364  & $\mathbf{\underline{0.829}}$  & $\mathbf{\underline{0.933}}$  & $\mathbf{\underline{0.964}}$  & 345  & 0.714  & 0.840  & 0.872  & $\mathbf{\underline{365}}$  & 0.280  & 0.730  & 0.922  & $\mathbf{\underline{365}}$  & 0.631  & 0.874  & 0.954 \\
0048 & 512 & $ 24.2\% $  & 501  & $\mathbf{\underline{0.802}}$  & $\mathbf{\underline{0.914}}$  & $\mathbf{\underline{0.949}}$  & 485  & 0.718  & 0.844  & 0.876  & $\mathbf{\underline{507}}$  & 0.389  & 0.769  & 0.930  & 505  & 0.679  & 0.883  & 0.948 \\
0024 & 356 & $ 23.0\% $  & 328  & $\mathbf{\underline{0.435}}$  & 0.636  & 0.702  & 308  & 0.391  & 0.620  & 0.709  & $\mathbf{\underline{355}}$  & 0.152  & 0.552  & $\mathbf{\underline{0.868}}$  & 338  & 0.326  & $\mathbf{\underline{0.648}}$  & 0.839 \\
0223 & 214 & $ 17.0\% $  & 211  & $\mathbf{\underline{0.582}}$  & $\mathbf{\underline{0.804}}$  & 0.890  & 202  & 0.430  & 0.673  & 0.770  & 212  & 0.014  & 0.335  & 0.768  & $\mathbf{\underline{213}}$  & 0.327  & 0.696  & $\mathbf{\underline{0.899}}$ \\
5016 & 28 & $ 16.9\% $  & $\mathbf{\underline{28}}$  & $\mathbf{\underline{0.790}}$  & $\mathbf{\underline{0.928}}$  & $\mathbf{\underline{0.984}}$  & $\mathbf{\underline{28}}$  & 0.355  & 0.499  & 0.741  & $\mathbf{\underline{28}}$  & 0.413  & 0.793  & 0.952  & $\mathbf{\underline{28}}$  & 0.508  & 0.835  & 0.960 \\
0046 & 440 & $ 14.6\% $  & 438  & $\mathbf{\underline{0.920}}$  & 0.962  & 0.976  & 388  & 0.472  & 0.534  & 0.573  & 434  & 0.515  & 0.837  & 0.939  & $\mathbf{\underline{440}}$  & 0.895  & $\mathbf{\underline{0.977}}$  & $\mathbf{\underline{0.996}}$ \\
\midrule \midrule
0099 & 299 & $ 47.4\% $  & 157  & 0.066  & 0.175  & 0.239  & 222  & 0.026  & 0.176  & 0.350  & $\mathbf{\underline{297}}$  & 0.011  & 0.169  & 0.610  & 255  & $\mathbf{\underline{0.382}}$  & $\mathbf{\underline{0.559}}$  & $\mathbf{\underline{0.661}}$ \\
1001 & 285 & $ 43.9\% $  & 268  & $\mathbf{\underline{0.016}}$  & $\mathbf{\underline{0.278}}$  & 0.616  & 256  & 0.004  & 0.194  & $\mathbf{\underline{0.628}}$  & $\mathbf{\underline{275}}$  & 0.000  & 0.001  & 0.047  & 270  & 0.000  & 0.003  & 0.119 \\
0231 & 296 & $ 42.2\% $  & 260  & $\mathbf{\underline{0.462}}$  & 0.669  & 0.747  & 247  & 0.377  & 0.595  & 0.664  & $\mathbf{\underline{286}}$  & 0.059  & 0.436  & 0.785  & 278  & 0.368  & $\mathbf{\underline{0.672}}$  & $\mathbf{\underline{0.813}}$ \\
0411 & 299 & $ 29.9\% $  & 268  & 0.556  & 0.731  & 0.785  & 273  & $\mathbf{\underline{0.578}}$  & $\mathbf{\underline{0.764}}$  & 0.820  & $\mathbf{\underline{293}}$  & 0.181  & 0.577  & $\mathbf{\underline{0.866}}$  & 268  & 0.305  & 0.584  & 0.745 \\
0377 & 295 & $ 27.5\% $  & 232  & $\mathbf{\underline{0.490}}$  & 0.583  & 0.609  & 197  & 0.342  & 0.418  & 0.440  & $\mathbf{\underline{269}}$  & 0.165  & 0.495  & 0.734  & 266  & 0.461  & $\mathbf{\underline{0.670}}$  & $\mathbf{\underline{0.765}}$ \\
0102 & 299 & $ 25.8\% $  & $\mathbf{\underline{296}}$  & $\mathbf{\underline{0.748}}$  & $\mathbf{\underline{0.904}}$  & $\mathbf{\underline{0.959}}$  & 280  & 0.595  & 0.794  & 0.857  & 294  & 0.163  & 0.459  & 0.759  & 293  & 0.525  & 0.773  & 0.908 \\
0147 & 298 & $ 24.6\% $  & $\mathbf{\underline{289}}$  & $\mathbf{\underline{0.597}}$  & $\mathbf{\underline{0.732}}$  & $\mathbf{\underline{0.869}}$  & 241  & 0.272  & 0.361  & 0.454  & 284  & 0.050  & 0.425  & 0.700  & $\mathbf{\underline{289}}$  & 0.000  & 0.000  & 0.000 \\
0148 & 287 & $ 24.6\% $  & 244  & $\mathbf{\underline{0.364}}$  & $\mathbf{\underline{0.504}}$  & $\mathbf{\underline{0.669}}$  & 256  & 0.331  & 0.434  & 0.534  & 275  & 0.045  & 0.257  & 0.458  & $\mathbf{\underline{282}}$  & 0.286  & 0.456  & 0.570 \\
0446 & 298 & $ 22.1\% $  & $\mathbf{\underline{296}}$  & $\mathbf{\underline{0.668}}$  & $\mathbf{\underline{0.884}}$  & $\mathbf{\underline{0.959}}$  & 288  & 0.586  & 0.832  & 0.909  & 289  & 0.050  & 0.433  & 0.790  & 295  & 0.456  & 0.783  & 0.928 \\
0022 & 297 & $ 21.2\% $  & 280  & $\mathbf{\underline{0.627}}$  & $\mathbf{\underline{0.822}}$  & 0.877  & 273  & 0.571  & 0.774  & 0.832  & $\mathbf{\underline{296}}$  & 0.193  & 0.627  & $\mathbf{\underline{0.894}}$  & 280  & 0.421  & 0.708  & 0.843 \\
0327 & 298 & $ 21.0\% $  & 280  & $\mathbf{\underline{0.677}}$  & $\mathbf{\underline{0.814}}$  & $\mathbf{\underline{0.860}}$  & 270  & 0.616  & 0.764  & 0.807  & $\mathbf{\underline{288}}$  & 0.038  & 0.503  & 0.826  & $\mathbf{\underline{288}}$  & 0.443  & 0.638  & 0.725 \\
0015 & 284 & $ 20.6\% $  & 241  & $\mathbf{\underline{0.552}}$  & 0.645  & 0.684  & 223  & 0.337  & 0.442  & 0.498  & 244  & 0.099  & 0.353  & 0.578  & $\mathbf{\underline{272}}$  & 0.525  & $\mathbf{\underline{0.769}}$  & $\mathbf{\underline{0.872}}$ \\
0455 & 298 & $ 19.8\% $  & 292  & $\mathbf{\underline{0.706}}$  & 0.856  & 0.923  & 293  & 0.688  & $\mathbf{\underline{0.885}}$  & 0.946  & 294  & 0.204  & 0.627  & 0.880  & $\mathbf{\underline{297}}$  & 0.581  & 0.856  & $\mathbf{\underline{0.960}}$ \\
0496 & 297 & $ 19.2\% $  & 280  & $\mathbf{\underline{0.641}}$  & $\mathbf{\underline{0.808}}$  & 0.858  & 281  & 0.550  & 0.784  & 0.862  & 285  & 0.074  & 0.458  & 0.781  & $\mathbf{\underline{291}}$  & 0.423  & 0.755  & $\mathbf{\underline{0.905}}$ \\
1589 & 299 & $ 17.4\% $  & 294  & 0.632  & 0.868  & 0.939  & 290  & $\mathbf{\underline{0.700}}$  & $\mathbf{\underline{0.875}}$  & 0.928  & 288  & 0.146  & 0.454  & 0.740  & $\mathbf{\underline{298}}$  & 0.579  & 0.806  & $\mathbf{\underline{0.945}}$ \\
0012 & 299 & $ 16.3\% $  & $\mathbf{\underline{295}}$  & $\mathbf{\underline{0.765}}$  & $\mathbf{\underline{0.883}}$  & 0.931  & 291  & 0.646  & 0.781  & 0.816  & 129  & 0.016  & 0.092  & 0.156  & $\mathbf{\underline{295}}$  & 0.627  & 0.863  & $\mathbf{\underline{0.945}}$ \\
0104 & 284 & $ 16.2\% $  & 237  & 0.411  & 0.486  & 0.589  & 195  & 0.342  & 0.439  & 0.465  & 265  & 0.110  & 0.374  & 0.559  & $\mathbf{\underline{280}}$  & $\mathbf{\underline{0.432}}$  & $\mathbf{\underline{0.616}}$  & $\mathbf{\underline{0.697}}$ \\
0019 & 299 & $ 15.4\% $  & 288  & $\mathbf{\underline{0.793}}$  & 0.881  & 0.916  & 247  & 0.567  & 0.656  & 0.677  & 271  & 0.200  & 0.533  & 0.729  & $\mathbf{\underline{296}}$  & 0.725  & $\mathbf{\underline{0.906}}$  & $\mathbf{\underline{0.966}}$ \\
0063 & 293 & $ 14.5\% $  & 266  & $\mathbf{\underline{0.620}}$  & 0.771  & 0.814  & 263  & 0.567  & 0.743  & 0.794  & 268  & 0.088  & 0.440  & 0.735  & $\mathbf{\underline{287}}$  & 0.441  & $\mathbf{\underline{0.773}}$  & $\mathbf{\underline{0.917}}$ \\
0130 & 285 & $ 14.4\% $  & 202  & $\mathbf{\underline{0.386}}$  & 0.457  & 0.493  & 193  & 0.327  & 0.417  & 0.444  & 187  & 0.025  & 0.182  & 0.356  & $\mathbf{\underline{279}}$  & 0.333  & $\mathbf{\underline{0.526}}$  & $\mathbf{\underline{0.800}}$ \\
0080 & 284 & $ 12.9\% $  & 141  & 0.186  & 0.230  & 0.242  & 140  & 0.172  & 0.220  & 0.235  & 278  & 0.033  & 0.267  & 0.690  & $\mathbf{\underline{282}}$  & $\mathbf{\underline{0.255}}$  & $\mathbf{\underline{0.505}}$  & $\mathbf{\underline{0.893}}$ \\
0240 & 298 & $ 11.9\% $  & 288  & $\mathbf{\underline{0.666}}$  & $\mathbf{\underline{0.856}}$  & $\mathbf{\underline{0.910}}$  & 276  & 0.514  & 0.718  & 0.791  & 285  & 0.129  & 0.490  & 0.792  & $\mathbf{\underline{290}}$  & 0.348  & 0.672  & 0.874 \\
0007 & 290 & $ 11.7\% $  & 284  & $\mathbf{\underline{0.759}}$  & 0.860  & 0.909  & 172  & 0.251  & 0.318  & 0.342  & 277  & 0.063  & 0.539  & 0.823  & $\mathbf{\underline{289}}$  & 0.655  & $\mathbf{\underline{0.875}}$  & $\mathbf{\underline{0.967}}$ \\
Mean & 379 & $ 26.1\% $  & 327  & $\mathbf{\underline{0.562}}$  & $\mathbf{\underline{0.703}}$  & 0.775  & 307  & 0.443  & 0.593  & 0.670  & 347  & 0.147  & 0.470  & 0.726  & $\mathbf{\underline{352}}$  & 0.443  & 0.671  & $\mathbf{\underline{0.799}}$ \\

    \end{tabular}
    \end{adjustbox}

\end{table*}

\vspace{-2mm}

  \begin{table*}[]
\caption{{\small {\bf 1DSfM experiment (registration-aware AUC).}
For each scene, we list the number of input images ($N_c$) and the fraction of outliers.
For each model, we report the number of registered cameras ($N_r$) and the registration-aware AUC (RA-AUC) values at different error thresholds (in degrees).
RA-AUC is computed over all image pairs; pairs involving at least one unregistered camera are assigned an error of $180^\circ$ and therefore contribute zero recall at all reported thresholds.
Winning results are marked in \textbf{\underline{bold and underlined}}.}}
      \label{tabapp:1dsfm_ra_auc}
      \centering
      \tiny
      \rowcolors{3}{rowgray}{white}
      \setlength{\tabcolsep}{3pt}
      \renewcommand{\arraystretch}{1.2}
      \begin{adjustbox}{max width=\textwidth}
      \begin{tabular}{lcc|cccc|cccc|cccc|cccc}

  \multirow{2}{*}{ Scene } &
  \multirow{2}{*}{ $N_c$ } &
  \multirow{2}{*}{ Outliers\% } &
  \multicolumn{4}{c|}{\textbf{Ours}}&
  \multicolumn{4}{c|}{RESFM} &
  \multicolumn{4}{c|}{\theia}&
  \multicolumn{4}{c}{\glomap}\\

   &  &   & $N_r$ & @1 & @5 & @30 & $N_r$ & @1 & @5 & @30 & $N_r$ & @1 & @5 & @30 & $N_r$ & @1 & @5 & @30 \\

Alamo & 573 & $ 32.6\% $  & 523  & $\mathbf{\underline{0.358}}$  & $\mathbf{\underline{0.559}}$  & 0.701  & 496  & 0.306  & 0.500  & 0.587  & 549  & 0.002  & 0.085  & 0.458  & $\mathbf{\underline{557}}$  & 0.087  & 0.456  & $\mathbf{\underline{0.787}}$ \\
Ellis Island & 227 & $ 25.1\% $  & 215  & $\mathbf{\underline{0.378}}$  & $\mathbf{\underline{0.754}}$  & $\mathbf{\underline{0.865}}$  & 211  & 0.246  & 0.662  & 0.801  & 213  & 0.000  & 0.020  & 0.388  & $\mathbf{\underline{219}}$  & 0.066  & 0.501  & 0.838 \\
Madrid Metropolis & 333 & $ 39.4\% $  & 298  & $\mathbf{\underline{0.472}}$  & $\mathbf{\underline{0.656}}$  & 0.756  & 247  & 0.188  & 0.308  & 0.385  & 319  & 0.013  & 0.221  & 0.596  & $\mathbf{\underline{320}}$  & 0.150  & 0.578  & $\mathbf{\underline{0.809}}$ \\
Montreal Notre Dame & 448 & $ 31.7\% $  & 442  & $\mathbf{\underline{0.490}}$  & $\mathbf{\underline{0.766}}$  & 0.870  & 333  & 0.291  & 0.436  & 0.515  & 420  & 0.001  & 0.082  & 0.486  & $\mathbf{\underline{444}}$  & 0.088  & 0.539  & $\mathbf{\underline{0.874}}$ \\
NYC Library & 330 & $ 33.6\% $  & 295  & $\mathbf{\underline{0.463}}$  & $\mathbf{\underline{0.681}}$  & 0.766  & 281  & 0.207  & 0.378  & 0.449  & 313  & 0.005  & 0.145  & 0.509  & $\mathbf{\underline{323}}$  & 0.136  & 0.608  & $\mathbf{\underline{0.872}}$ \\
Notre Dame & 549 & $ 35.6\% $  & 527  & $\mathbf{\underline{0.393}}$  & $\mathbf{\underline{0.714}}$  & $\mathbf{\underline{0.876}}$  & 510  & 0.326  & 0.629  & 0.792  & 531  & 0.014  & 0.274  & 0.679  & $\mathbf{\underline{543}}$  & 0.099  & 0.553  & 0.846 \\
Piazza del Popolo & 336 & $ 33.1\% $  & 318  & $\mathbf{\underline{0.390}}$  & 0.539  & 0.678  & 248  & 0.323  & 0.436  & 0.486  & 324  & 0.023  & 0.210  & 0.565  & $\mathbf{\underline{331}}$  & 0.197  & $\mathbf{\underline{0.629}}$  & $\mathbf{\underline{0.873}}$ \\
Tower of London & 467 & $ 27.0\% $  & 457  & $\mathbf{\underline{0.466}}$  & $\mathbf{\underline{0.785}}$  & $\mathbf{\underline{0.905}}$  & 116  & 0.015  & 0.035  & 0.044  & 448  & 0.002  & 0.085  & 0.436  & $\mathbf{\underline{466}}$  & 0.114  & 0.597  & 0.893 \\
Vienna Cathedral & 824 & $ 31.4\% $  & 763  & $\mathbf{\underline{0.318}}$  & $\mathbf{\underline{0.502}}$  & 0.608  & 578  & 0.167  & 0.287  & 0.347  & 767  & 0.000  & 0.007  & 0.234  & $\mathbf{\underline{822}}$  & 0.052  & 0.496  & $\mathbf{\underline{0.842}}$ \\
Yorkminster & 432 & $ 29.0\% $  & 402  & $\mathbf{\underline{0.443}}$  & $\mathbf{\underline{0.735}}$  & 0.833  & 304  & 0.040  & 0.200  & 0.311  & 387  & 0.000  & 0.030  & 0.287  & $\mathbf{\underline{418}}$  & 0.097  & 0.561  & $\mathbf{\underline{0.847}}$ \\
Mean & 451 & $ 31.9\% $  & 424  & $\mathbf{\underline{0.417}}$  & $\mathbf{\underline{0.669}}$  & 0.786  & 332  & 0.211  & 0.387  & 0.472  & 427  & 0.006  & 0.116  & 0.464  & $\mathbf{\underline{444}}$  & 0.109  & 0.552  & $\mathbf{\underline{0.848}}$ \\

    \end{tabular}
    \end{adjustbox}

\end{table*}

\end{document}